\documentclass[letterpaper, 10 pt, conference]{ieeeconf}  
                                                          
\IEEEoverridecommandlockouts                              
                                                          
\newcommand{\CO}[1]{}
                                                          
\usepackage[ruled,vlined]{algorithm2e}
\usepackage{algorithmic}

\usepackage{url}
\usepackage{cite}
\usepackage{graphicx}
\usepackage{epsfig}
\usepackage{latexsym}
\usepackage{epic,eepic,color}
\usepackage{times}
\usepackage{mathptmx}

\usepackage{times}
\usepackage{multirow}

\definecolor{red}{rgb}{1,0,0}
\definecolor{green}{rgb}{0,1,0}
\definecolor{blue}{rgb}{0,0,1}
\definecolor{violet}{rgb}{1,0,1}
\definecolor{cyan}{cmyk}{1,0,0,0}
\definecolor{magenta}{cmyk}{0,1,0,0}
\definecolor{yellow}{cmyk}{0,0,1,0}

\newcommand{\4}{\color{blue}}
\newcommand{\5}{\color{blue}\bf}

\newcommand{\CommentOut}[1]{}

\newcommand{\FIG}[3]{
\begin{minipage}[b]{#1cm}
\begin{center}
\includegraphics[width=#1cm]{#2}
{\scriptsize #3}
\end{center}
\end{minipage}
}

\newcommand{\FIGR}[3]{
\begin{minipage}[b]{#1cm}
\begin{center}
\includegraphics[angle=-90,clip,width=#1cm]{#2}\vspace*{1mm}
\\
{\scriptsize #3}
\vspace*{1mm}
\end{center}
\end{minipage}
}

\newcommand{\FIGpng}[5]{
\begin{minipage}[b]{#1cm}
\begin{center}
\includegraphics[bb=0 0 #4 #5, clip, width=#1cm]{#2}\vspace*{-1mm}
\\
{\scriptsize #3}
\vspace*{1mm}
\end{center}
\end{minipage}
}

\newcommand{\FIGpnga}[3]{\FIGpng{#1}{#2}{#3}{640}{640}}

\title{\LARGE \bf 
Discriminative Map Retrieval Using View-Dependent Map Descriptor
}

\author{Liu Enfu ~~~~~~~~~~~~~~~~ Tanaka Kanji
\thanks{Our work has been supported in part by 
JSPS KAKENHI 
Grant-in-Aid for Young Scientists (B) 23700229,
and for Scientific Research (C) 26330297.}
\thanks{E. Liu and K. Tanaka are with Graduate School of Engineering, University of Fukui, Japan. 
{\tt\small tnkknj@u-fukui.ac.jp}}%
\vspace*{-5mm}}

\begin{document}

\maketitle
\thispagestyle{empty}
\pagestyle{empty}

\newcommand{\figBca}{
\begin{figure}[t]
\begin{center}
\FIG{2.5}{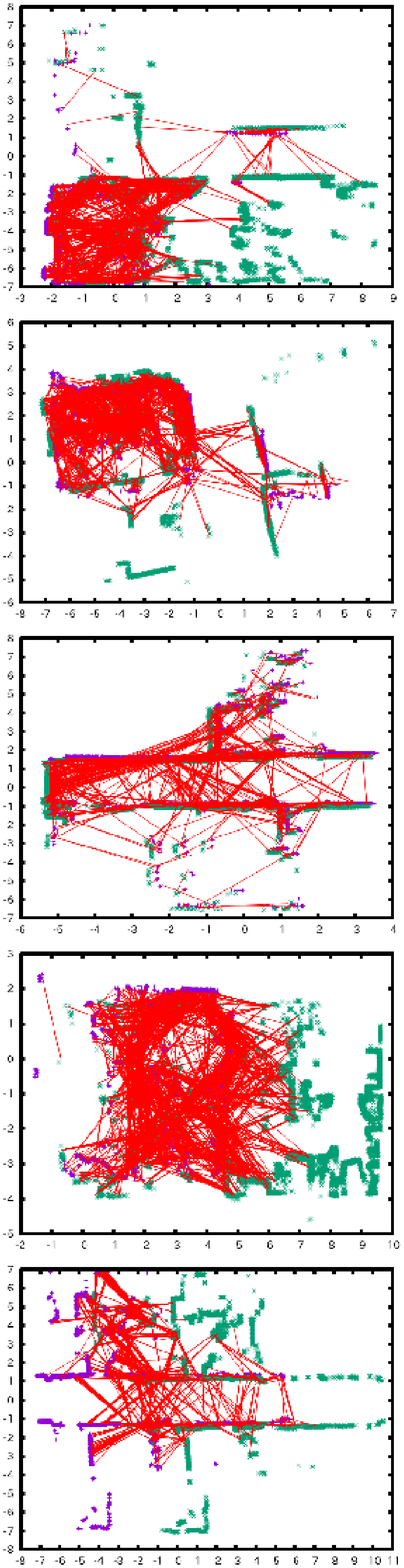}{$S^5$}\hspace*{-2mm}%
\FIG{2.5}{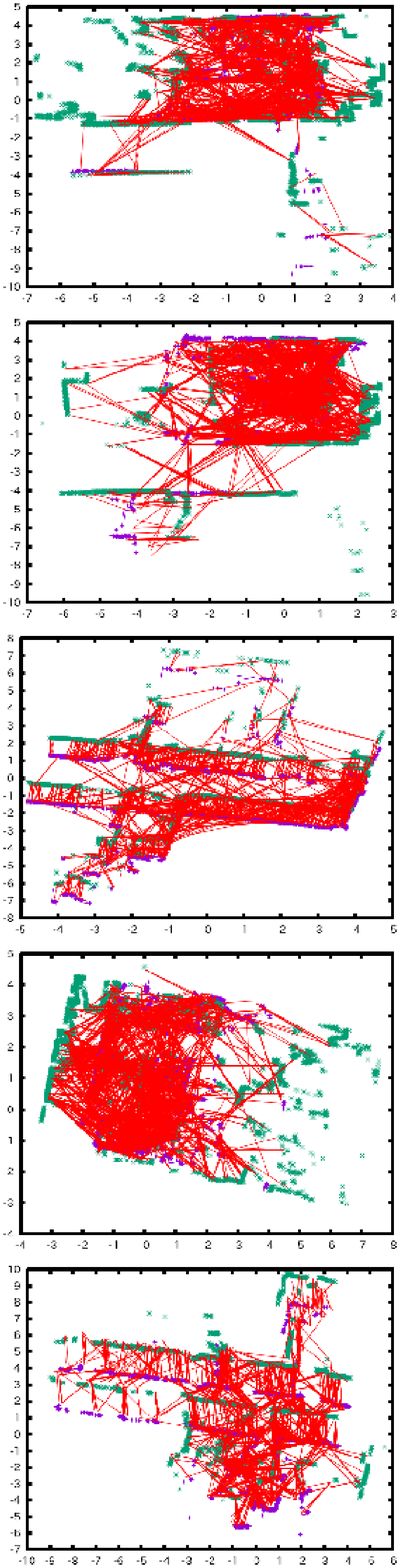}{$S^1$}\hspace*{-2mm}%
\FIG{2.5}{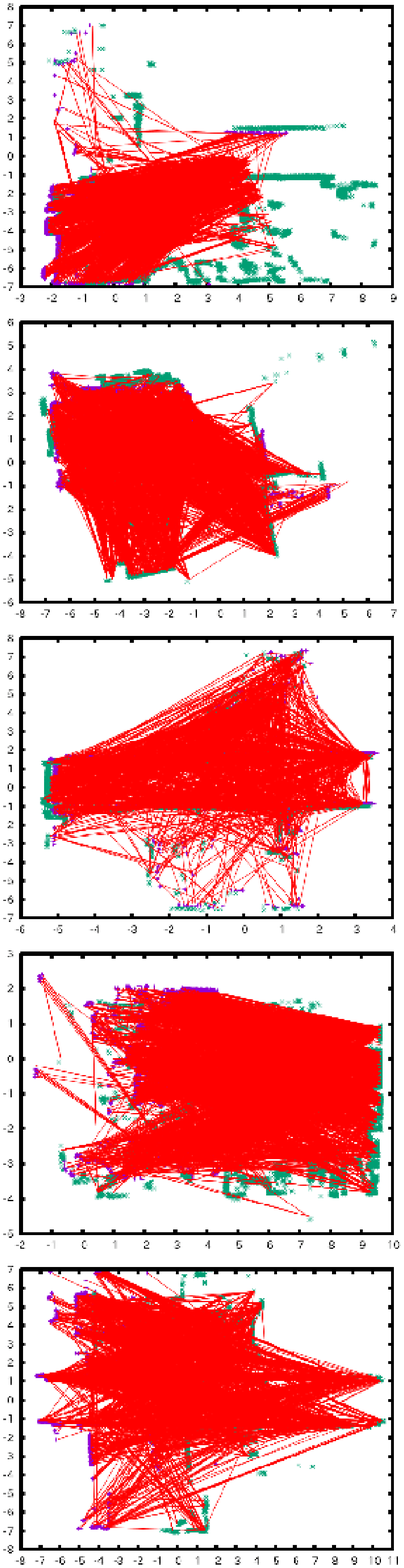}{BoW}\vspace*{-3mm}\\
\caption{Examples of matching visual words between relevant map pairs. Red lines
connect matched visual words between query and relevant database maps.}\label{fig:Bbca}
\end{center}
\vspace*{-5mm}
\end{figure}
}

\newcommand{\figBcb}{
\begin{figure}[t]
\begin{center}
\FIG{2.5}{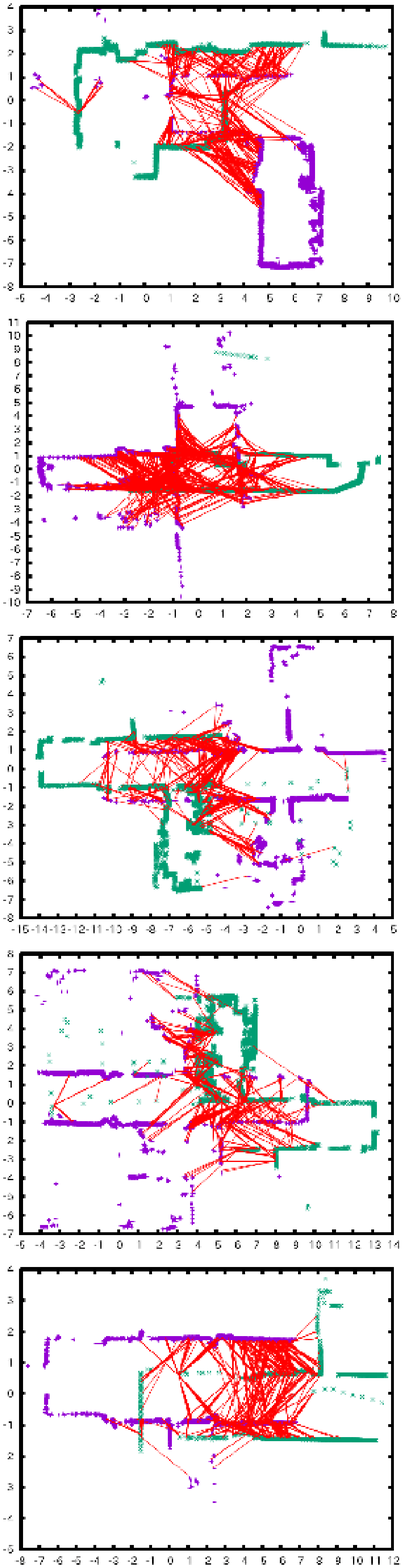}{$S^5$}\hspace*{-2mm}%
\FIG{2.5}{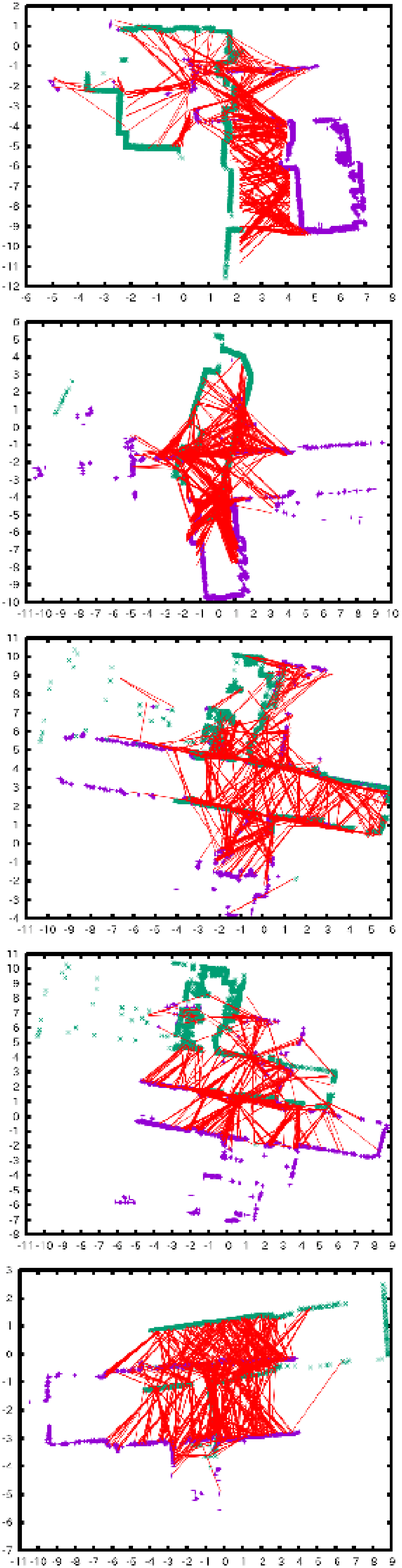}{$S^1$}\hspace*{-2mm}%
\FIG{2.5}{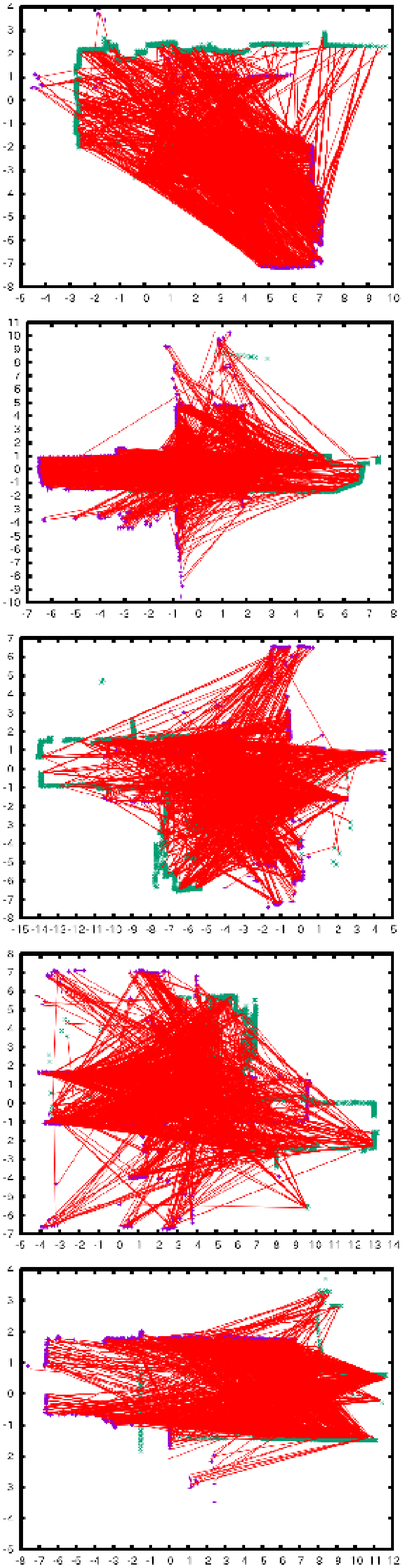}{BoW}\vspace*{-3mm}\\
\caption{Examples of matching visual words between irrelevant map pairs.}\label{fig:Bbcb}
\end{center}
\end{figure}
}

\newcommand{\figD}{
\begin{figure}[t]
\begin{center}
\hspace*{5mm}\FIG{8}{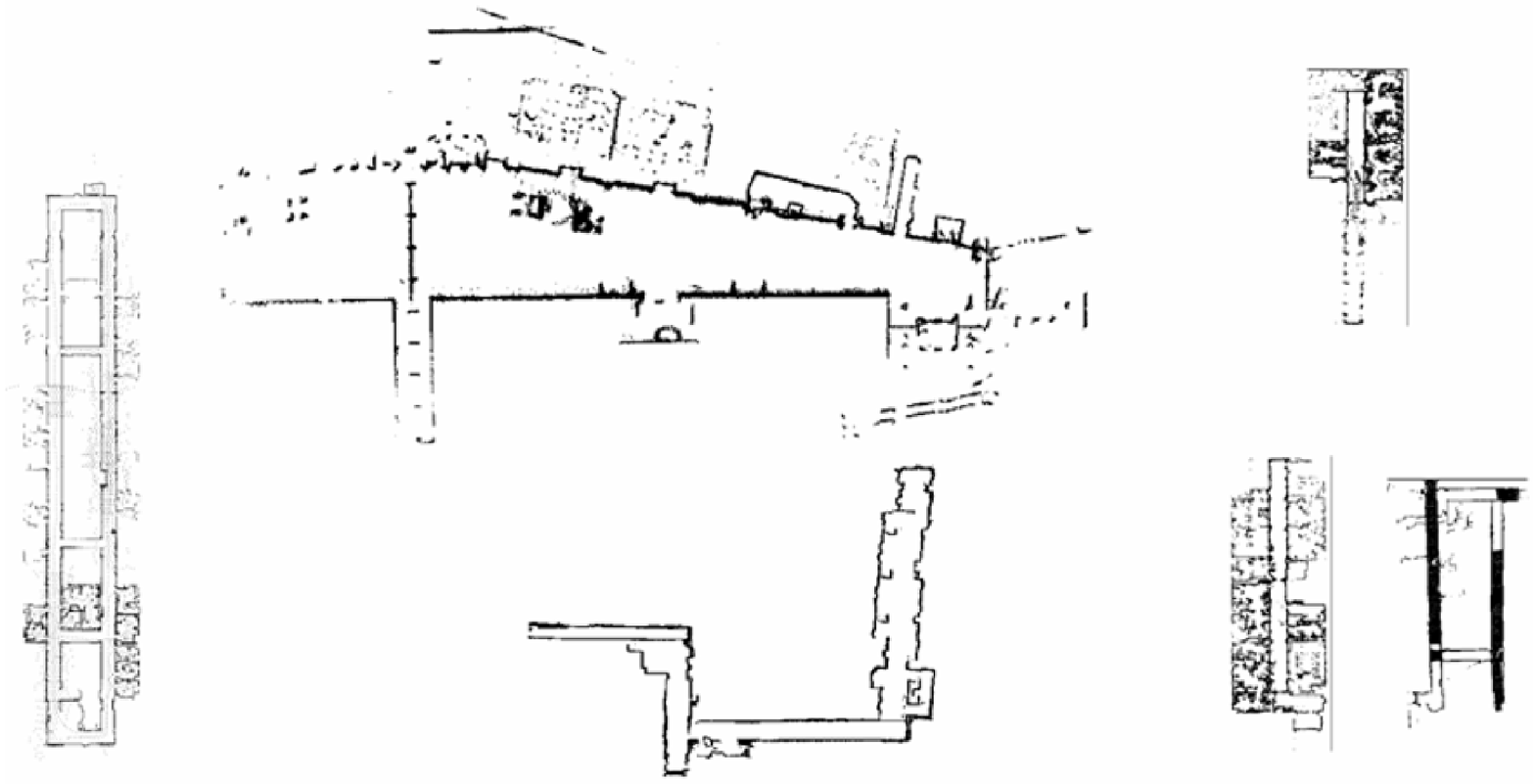}{}\\
\caption{Datasets used in the experiments: ``albert," ``fr079," ``run1," ``fr101," ``claxton," and ``kwing" from the radish dataset \cite{radish}.}\label{fig:D}
\end{center}
\vspace*{-3mm}
\end{figure}
}

\newcommand{\figM}{
\begin{figure}[t]
\begin{center}
\FIG{8}{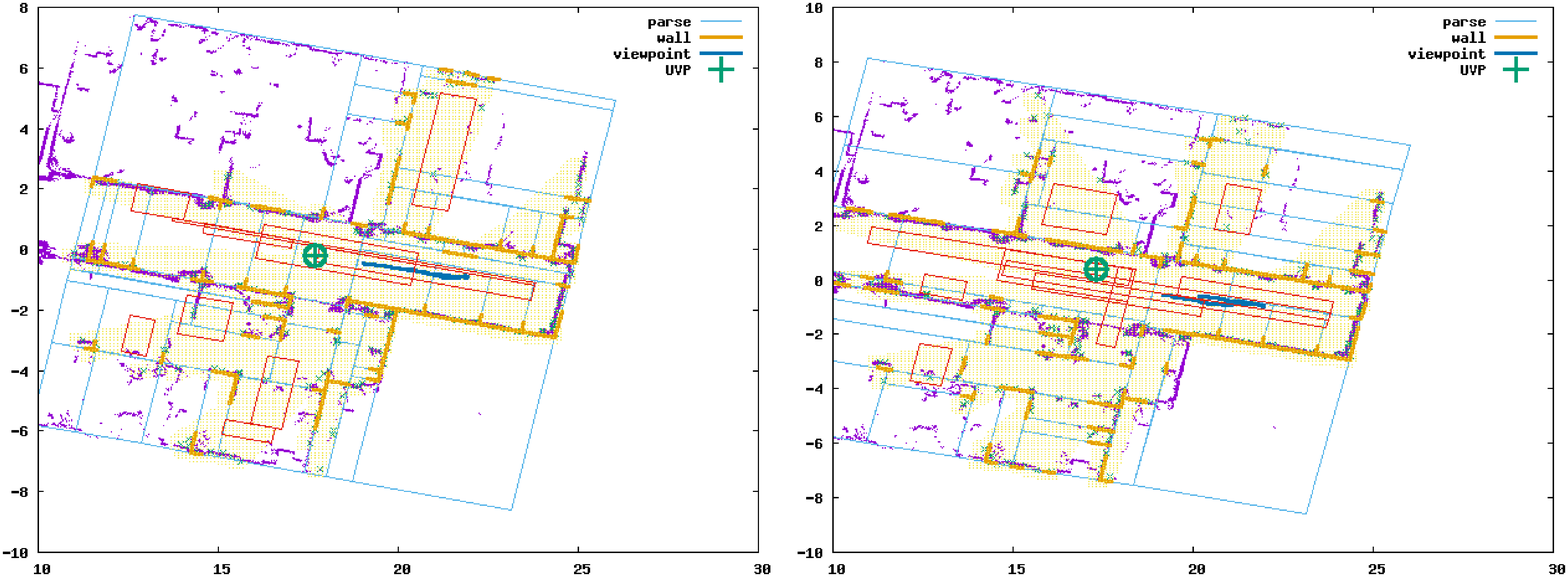}{a}\\
\FIG{8}{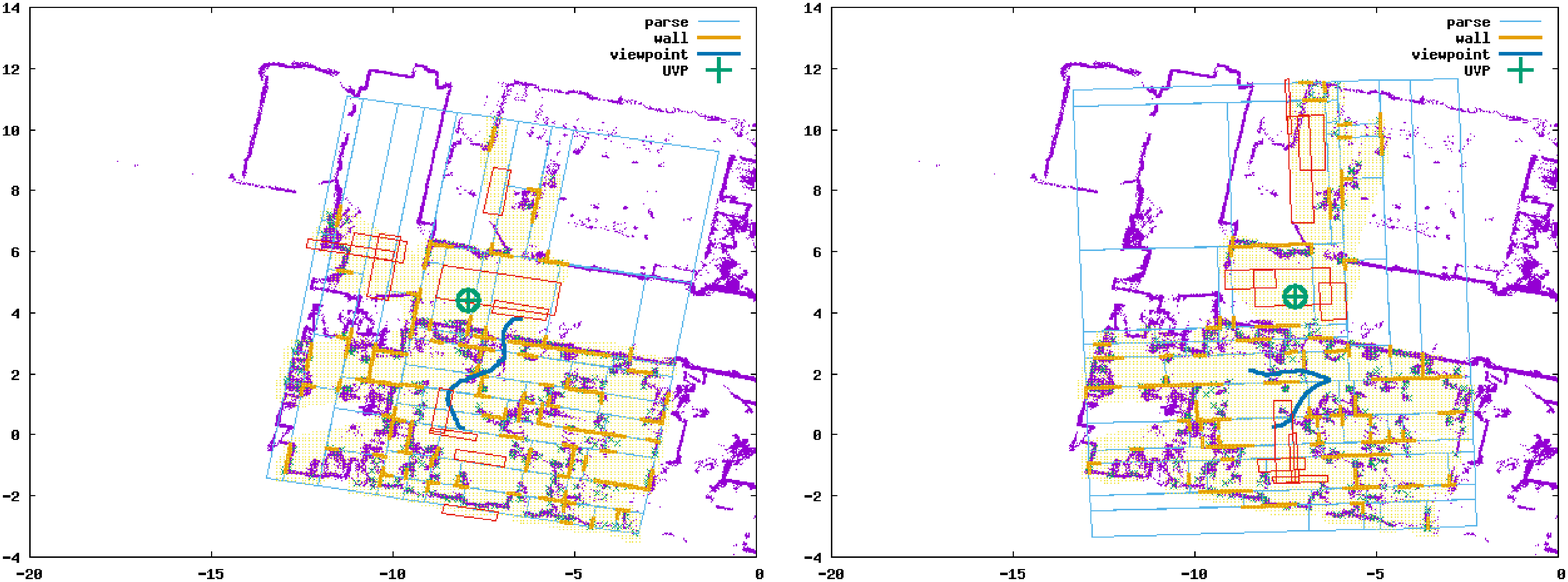}{b}\\
\FIG{8}{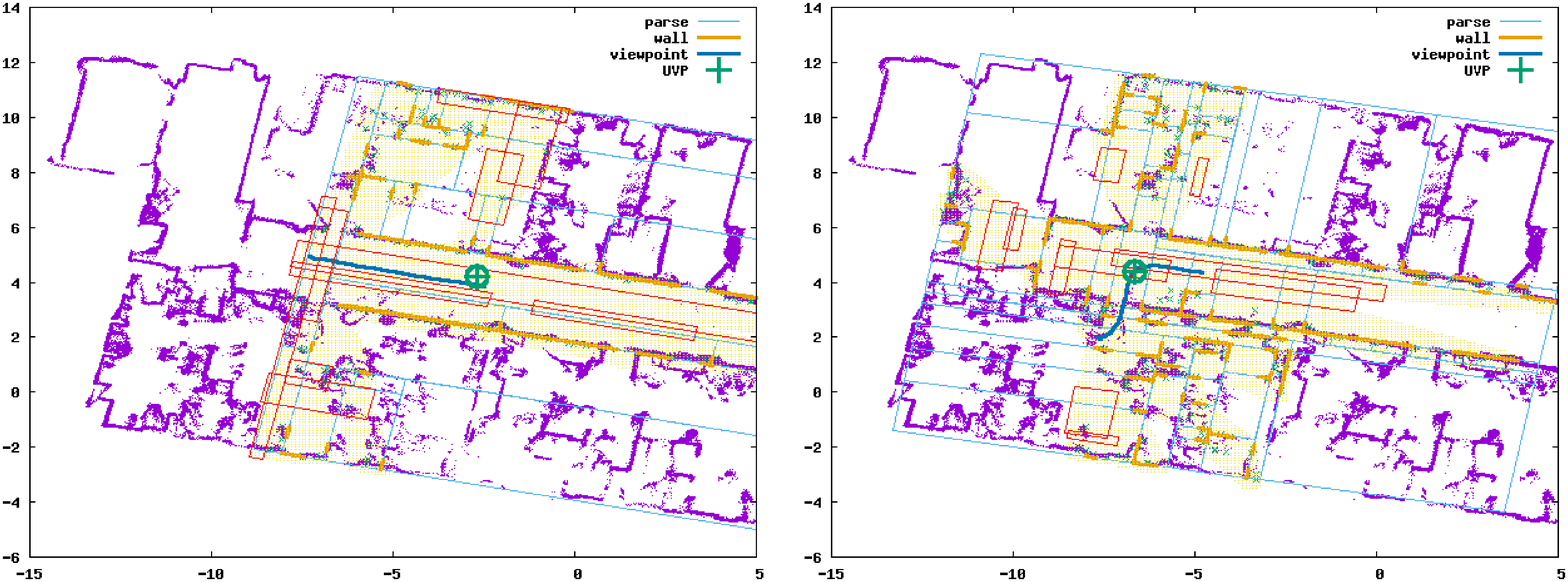}{c}\vspace*{-3mm}\\
\caption{Examples of viewpoint planning. Each panel shows the scene parsing for the query
map (left) and the relevant database map (right). We performed scene parsing using Manhattan
world grammar and obtained ``wall" primitives and unoccupied regions, as shown in the
figures. The red rectangles are the bounding boxes generated by strategy $S^5$.}\label{fig:M}
\end{center}
\end{figure}
}

\newcommand{\figN}{
\begin{figure}[t]
\begin{center}
\FIG{8.5}{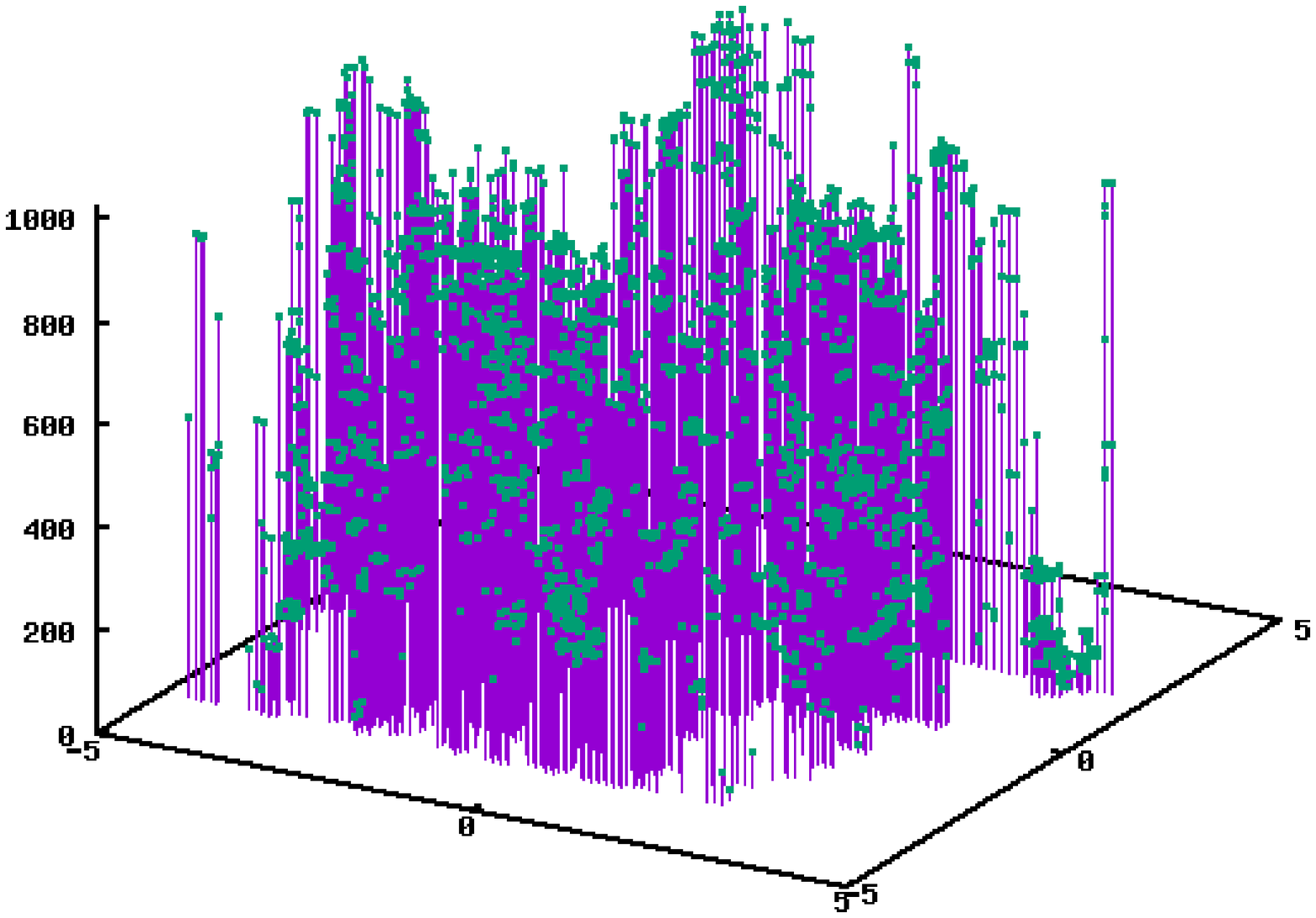}{}\\
\FIG{8.5}{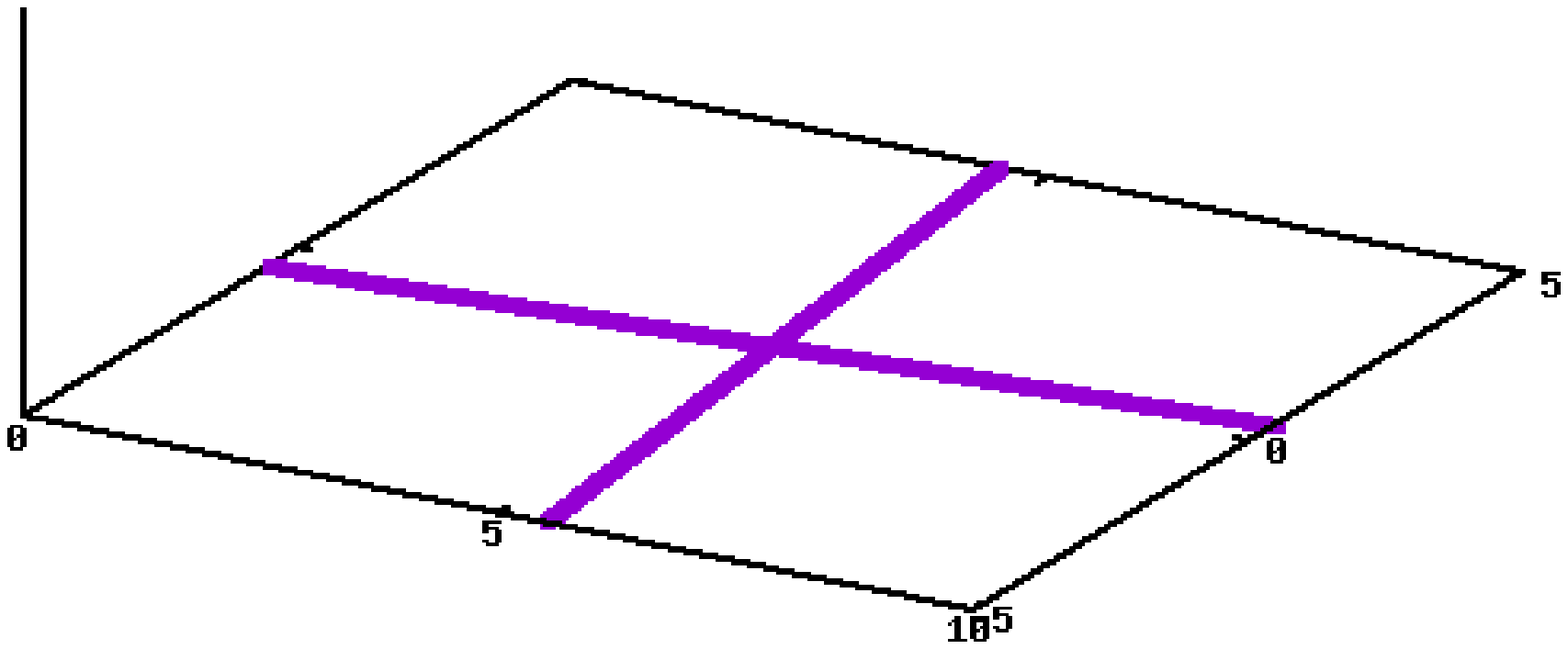}{}\\
\FIG{8.5}{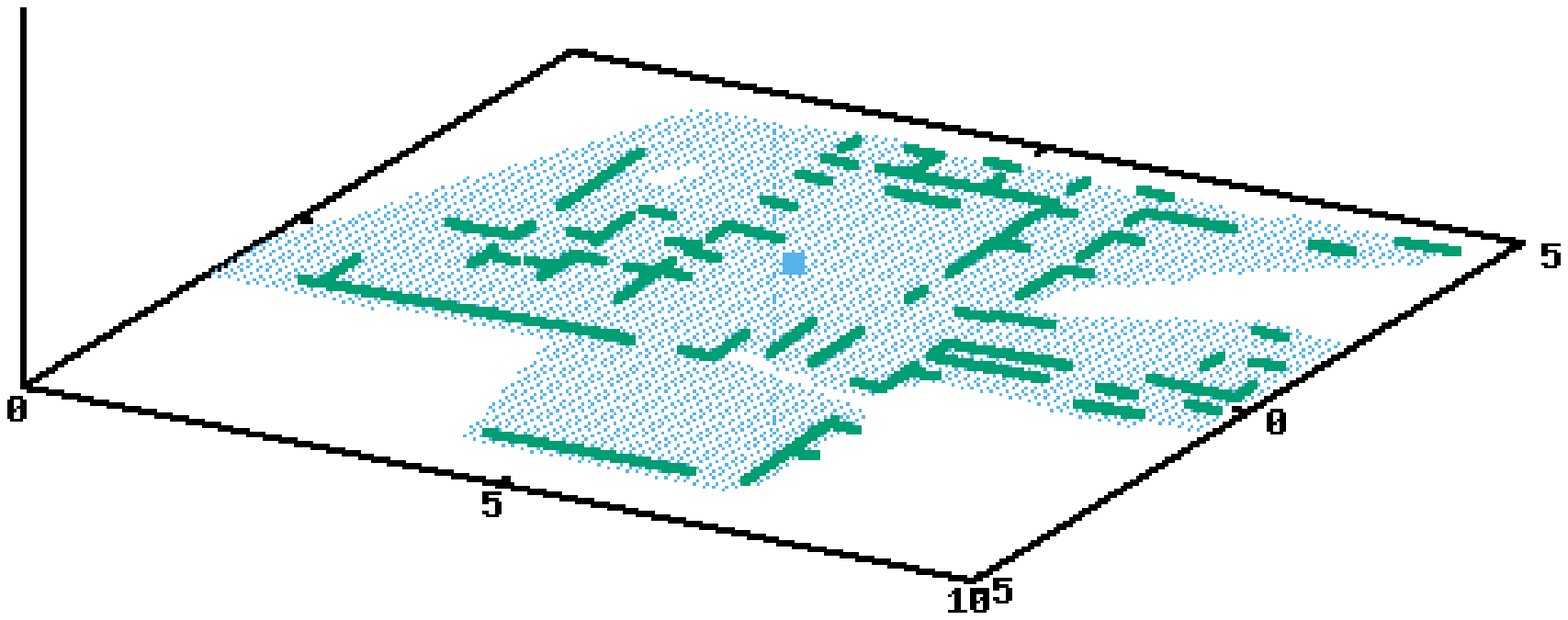}{}\\
\FIG{8.5}{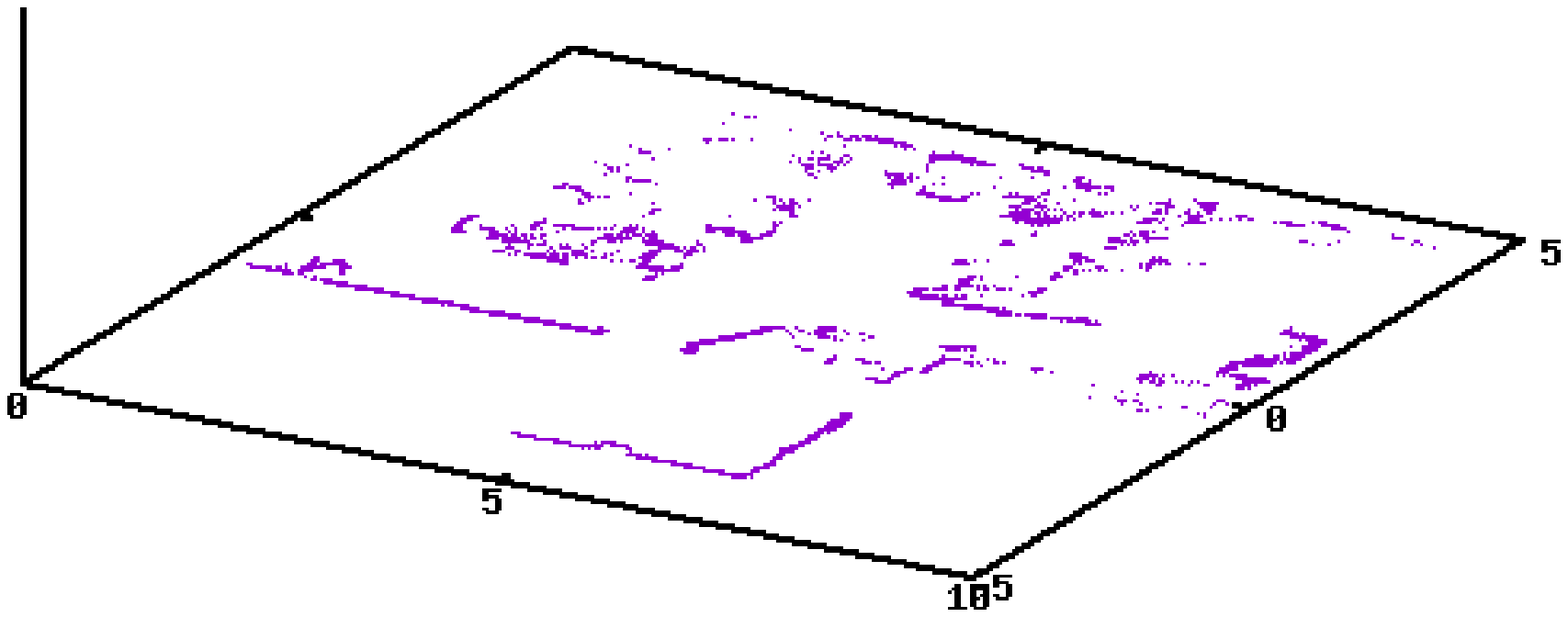}{}\vspace*{-1mm}\\
\caption{Local map descriptor (LMD). Unlike previous bag-of-words methods which
ignore all information about the layout of local features, we develop a holistic
descriptor that is view-dependent and highly discriminative. Our strategy is to explicitly model a unique viewpoint of an input local map; the pose of the local feature is defined with respect to this unique viewpoint, and can be viewed as an additional invariant feature for discriminative map retrieval. In the figure, from bottom to top, we observe an input pointset map, a result of scene parsing and viewpoint planning, a viewpoint-centric coordinate, and a set of visual words. Each visual word consists of an appearance word
$w_a$ (vertical axis)
and a pose word $\langle w_x, w_y \rangle$ 
(horizontal axes), which is defined with respect to the planned viewpoint and viewing direction.}\label{fig:N}
\vspace*{-7mm}
\end{center}
\end{figure}
}

\newcommand{\figO}{
\begin{figure}[t]
\begin{center}
\FIG{8}{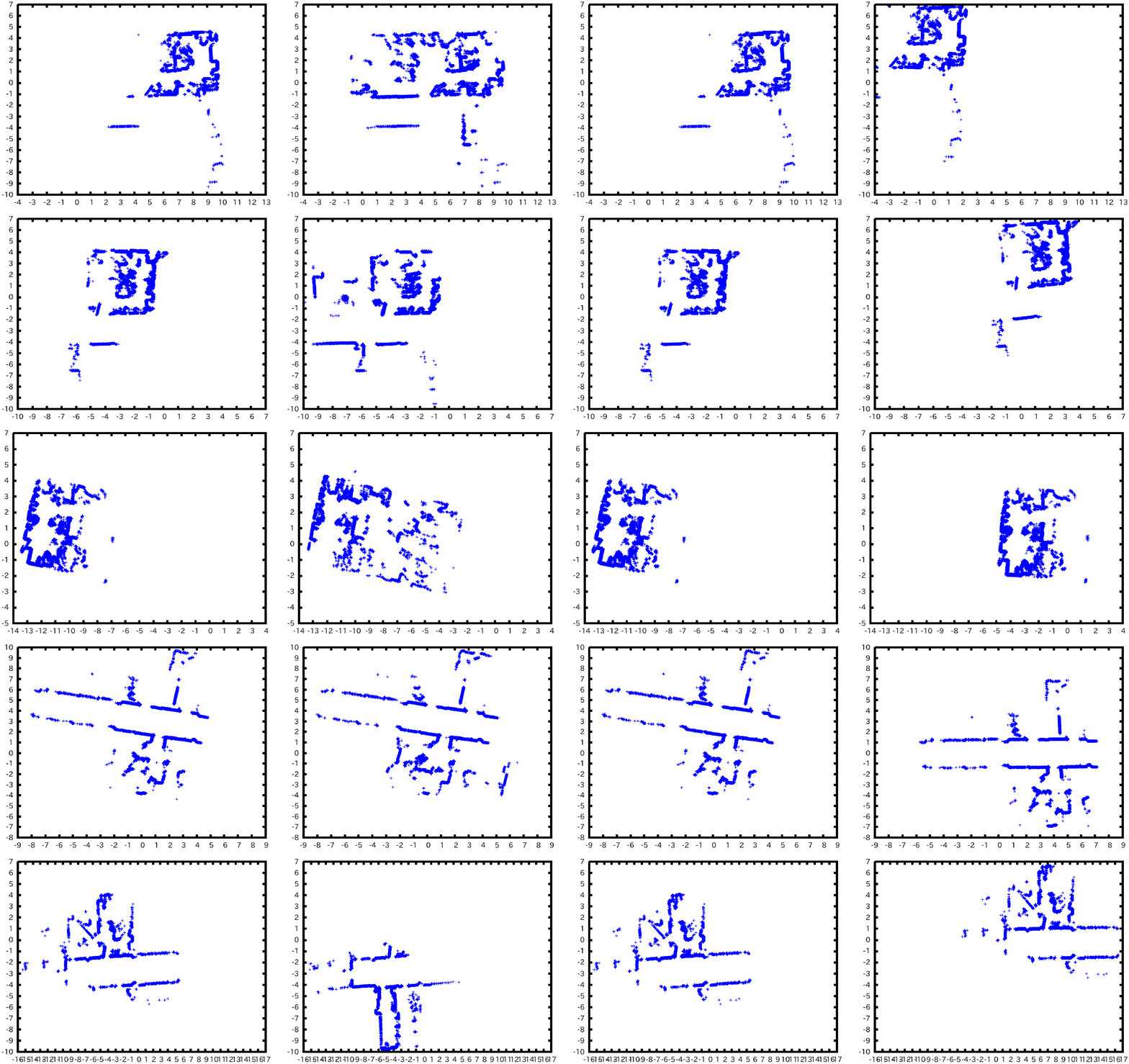}{}\vspace*{-3mm}\\
\caption{Examples of map retrieval. From left to right, a query map, the ground-truth database map, map retrieved by BoW method, and map retrieved by strategy $S^5$.}\label{fig:O}
\end{center}
\vspace*{-5mm}
\end{figure}
}

\newcommand{\figQ}{
\begin{figure}[t]
\begin{center}
\FIG{8}{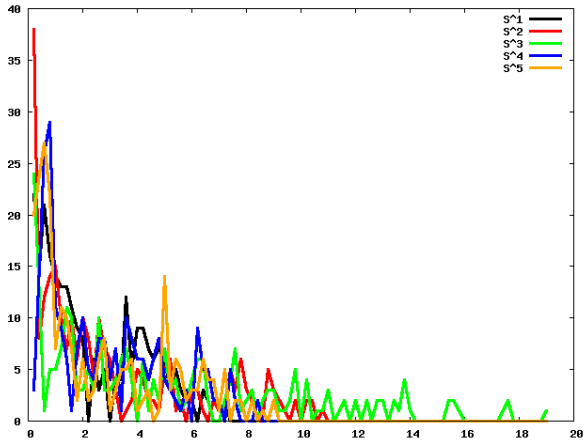}{}
\caption{Histogram of errors in viewpoint planning.}\label{fig:Q}
\end{center}
\vspace*{-5mm}
\end{figure}
}

\newcommand{\figR}{
\begin{figure}[t]
\vspace*{-8mm}
\begin{center}
\FIGR{6}{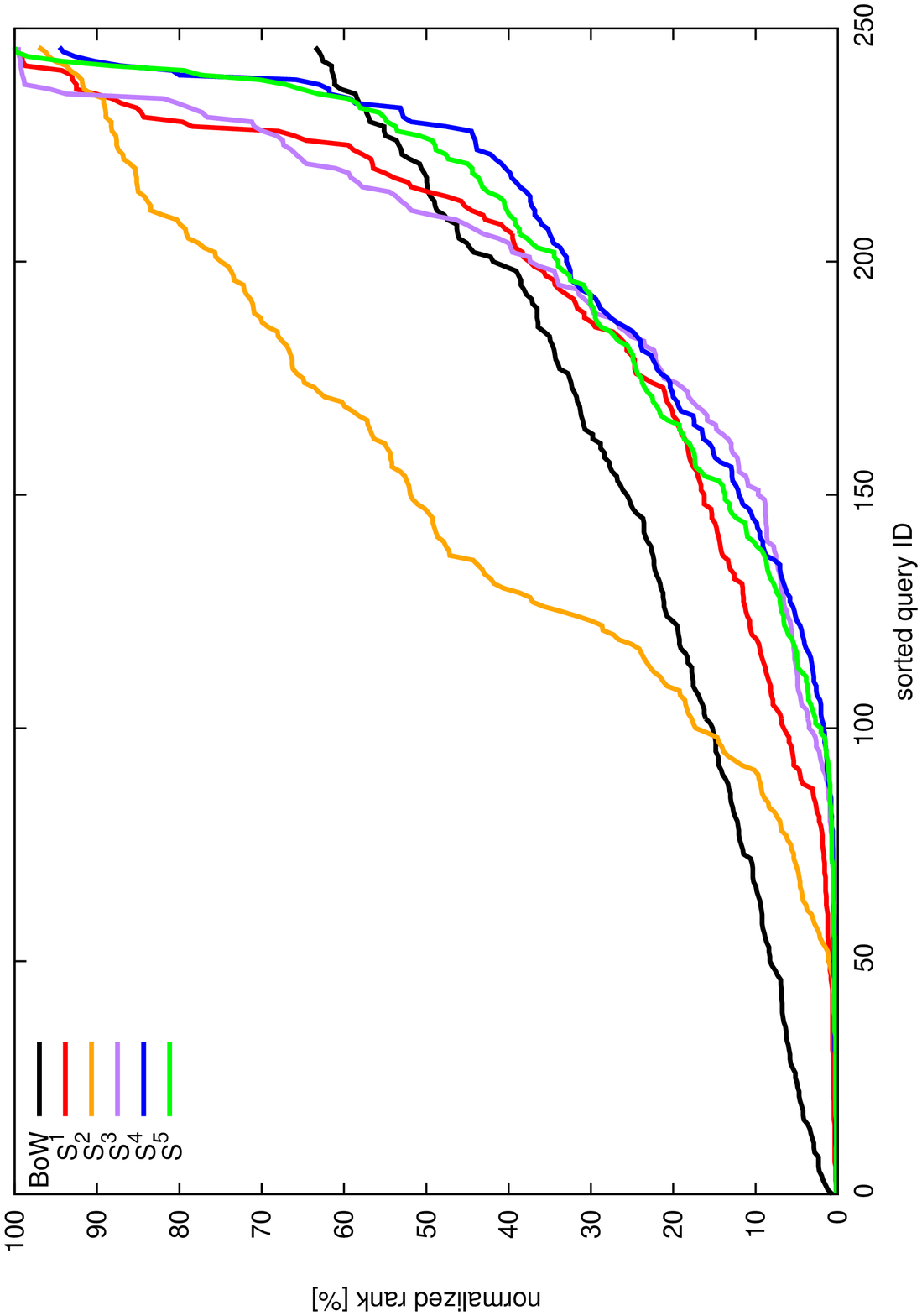}{}\vspace*{-5mm}\\
\caption{Performance in normalized rank [\%].}\label{fig:R}
\end{center}
\vspace*{-5mm}
\end{figure}
}

\newcommand{\tabB}{
\begin{table}[t]
\caption{Performance for dissimilar map pairs in ANR [\%]}\label{tab:B}
\label{table1}
\begin{center}
\vspace*{-5mm}
\begin{tabular}{|c|rrrrr|r|}
\hline
dataset & albert & fr079 & fr101 & kwing1 & run1 & Avg \\
\hline
\hline
BoW & 24.6 & 22.6 & 21.7 & 55.8 & 36.6& 25.4 \\
\hline
$S^1$ & 12.3 & 10.3 & 18.9 & 31.6 & 30.0 &14.4\\
\hline
$S^2$ & 20.7 & 11.3 & 28.8 & 29.3 & 26.3 & 19.4\\
\hline
$S^3$ & 33.6 & 34.8 & 39.8 & 45.0 & 51.2 & 36.3\\
\hline
$S^4$ &13.8&11.2 & 23.0 & 42.2 & 41.7 &16.9\\
\hline
$S^5$ & 15.7 & 14.5 & 17.4 & 32.3 & 36.6 & 16.9\\
\hline
\end{tabular}
\end{center}
\vspace*{-5mm}
\end{table}
}

\newcommand{\tabAd}{
\begin{table}[t]
\caption{Summary of ANR Performance [\%]}\label{tab:A}
\begin{center}
\vspace*{-5mm}
\begin{tabular}{|c|rrrrr|r|}
\hline
dataset & albert & fr079 & fr101 & kwing1 & run1 & Avg \\
\hline
\hline
BoW & 26.3 & 21.4 & 25.0 & 32.3 & 33.8& 24.8 \\
\hline
$S^1$ & 23.3 & 18.1 & {\4 21.5} & {\5 17.8} & {\5 19.0} &20.2\\
\hline
$S^2$ & 24.9 & {\5 9.4} & 28.6 & 27.0 & 27.7 & 18.6\\
\hline
$S^3$ & 32.3 & 37.2 & 41.3 & 46.5 & 47.6 & 37.4\\
\hline
$S^4$ & {\5 15.2} & {\4 10.0} & 23.8 & 21.5 & 35.9 & {\5 15.5}\\
\hline
$S^5$ & {\4 17.6} & 13.1 & {\5 20.4} & 21.4 & 34.8 & {\4 16.6}\\
\hline
\end{tabular}
\end{center}
\end{table}
}

\newcommand{\tabC}{
\begin{table}[t]
\caption{Performance for different parameter $D_{x,y}$ [\%]}\label{tab:B}
\begin{center}
\vspace*{-5mm}
\begin{tabular}{|c|rrrrr|r|}
\hline
dataset & albert & fr079 & fr101 & kwing1 & run1 & Avg \\
\hline
\hline
BoW & 26.3 & 21.4 & 25.0 & 32.3 & 33.7 & 24.8 \\ \hline
$S^1 (D_{x,y}=1[m])$ & 38.0 & 21.3 & 32.4 & 8.1 & 13.7 & 27.2 \\ \hline
$S^1 (D_{x,y}=5[m])$ & 25.4 & 13.1 & 15.1 & 33.1 & 22.3 & 17.8 \\ \hline
$S^5 (D_{x,y}=1[m])$ & 16.2 & 14.4 & 23.6 & 43.9 & 31.2 & 18.5 \\ \hline
$S^5 (D_{x,y}=5[m])$ & 20.7 & 13.7 & 21.3 & 41.2 & 34.5 & 18.8 \\ \hline
\end{tabular}
\end{center}
\end{table}
}

\newcommand{\ms}{\hspace*{-2mm}} % minus space

\newcommand{\figGd}{
\begin{figure}[t]
\begin{center}
\FIGpng{8}{merged.eps}{}
\caption{Own collected map dataset. 
From top to bottom, 
maps \#0-\#4, \#5-\#8, \#9-\#11, \#12-\#14, \#15-\#18, \#19-\#24, \#25-\#30, \#31-\#34, and \#35-\#38.
Rankings output by M2T and BoW methods are shown in each panel.
The details are better seen by zooming on a computer screen.}\label{fig:G}
\vspace*{-5mm}
\end{center}
\end{figure}
}

\newcommand{\figH}{
\begin{figure}[t]
\begin{center}
\FIGR{6}{rank.eps}{}\vspace*{-5mm}\\
\caption{Ranks for each query from our own collected dataset. (horizontal axis: query map ID, vertical axis: ANR in [\%]).}
\end{center}
\vspace*{-5mm}
\end{figure}
}

\newcommand{\figi}{
\begin{algorithm}[h]
{\small
\DontPrintSemicolon
\SetKwData{Left}{left}
\SetKwData{This}{this}
\SetKwData{Up}{up}
\SetKwFunction{VerticalSplit}{VerticalSplit}
\SetKwFunction{HorizontalSplit}{HorizontalSplit}
\SetKwFunction{SplitToWalls}{SplitToWalls}
\SetKwFunction{ScoreWall}{ScoreWall}
\SetKwFunction{DominantOrientation}{DominantOrientation}
\SetKwInOut{Input}{input}
\SetKwInOut{Output}{output}
\SetAlgoLined
%\TitleOfAlgo{Generate and Score Hypothesis}
\Input{point cloud $O$ and dominant orientation $\theta$.}
\Output{hypothesis and its score $h=(r_1, \cdots, r_N)$, $s$.}
\caption{HypothesizePolicy}
\label{alg1}
Initialize the Manhattan world $R_1$.\;
\For{$i=1$ to $N$}{
Sample ID of room to split $i$.\;
Sample splitting direction $d$.\;
\Switch{$d$}{
\uCase{``vertical"}{$R_i,R_N \Leftarrow$ \VerticalSplit($R_i$, $O$).\;}
\uCase{``horizontal"}{$R_i,R_N \Leftarrow$ \HorizontalSplit($R_i$, $O$).\;}
}}
$W \Leftarrow \{\}$.\;
\For{$i=1$ to $N$}{
$W \Leftarrow W\cup$ \SplitToWalls($R_i$).\;
}
$s \Leftarrow 0$.\;
\ForAll{$w\in W$}{
$s \Leftarrow s +$ \ScoreWall($w$, $O$).\;
}
}
\end{algorithm}
}

\newcommand{\figj}{
\begin{algorithm}[h]
{\small
\DontPrintSemicolon
\SetKwFunction{HypothesizeVerticalSplit}{HypothesizeVerticalSplit}
\SetKwFunction{SplitToRooms}{SplitToRooms}
\SetKwInOut{Input}{input}
\SetKwInOut{Output}{output}
\SetAlgoLined
%\TitleOfAlgo{Generate and Score Hypothesis}
\Input{room $R$.}
\Output{rooms $R$, $R'$.}
\caption{VerticalSplit}
\label{alg2}
$v^{best} \Leftarrow Null$; $s^{best} \Leftarrow -\infty$.\;
\For{$i=1$ to $H$}{
$v \Leftarrow$ \HypothesizeVerticalSplit($R$).\;
$s \Leftarrow $ \ScoreWall($v$, $O$).\;
\lIf{$s>s^{best}$}{
$v^{best} \Leftarrow$ v; $s^{best} \Leftarrow s.$\;
}
}
$R,R'\Leftarrow $ \SplitToRooms($v^{best}$, $R$).\;
}
\end{algorithm}
}

\newcommand{\figk}{
\begin{algorithm}[h]
{\small
\DontPrintSemicolon
\SetKwFunction{HypothesizeHorizontalSplit}{HypothesizeHorizontalSplit}
\SetKwFunction{SplitToRooms}{SplitToRooms}
\SetKwInOut{Input}{input}
\SetKwInOut{Output}{output}
\SetAlgoLined
%\TitleOfAlgo{Generate and Score Hypothesis}
\Input{room $R$.}
\Output{rooms $R$, $R'$.}
\caption{HorizontalSplit}
\label{alg3}
$h^{best} \Leftarrow Null$; $s^{best} \Leftarrow -\infty$.\;
\For{$i=1$ to $H$}{
$h \Leftarrow$ \HypothesizeHorizontalSplit($R$).\;
$s \Leftarrow $ \ScoreWall($h$, $O$).\;
\lIf{$s>s^{best}$}{
$h^{best} \Leftarrow$ h; $s^{best} \Leftarrow s.$\;
}
}
$R,R'\Leftarrow $ \SplitToRooms($h^{best}$, $R$).\;
}
\end{algorithm}
}

\newcommand{\figl}{
\begin{algorithm}[h]
{\small
\DontPrintSemicolon
\SetKwFunction{DominantOrientation}{DominantOrientation}
\SetKwFunction{HypothesizePolicy}{HypothesizePolicy}
\SetKwInOut{Input}{input}
\SetKwInOut{Output}{output}
\SetAlgoLined
%\TitleOfAlgo{Generate and Score Hypothesis}
\Input{point cloud $O$.}
\Output{best policy $p^{best}$.}
\caption{MapParsing}
\label{alg4}
$\theta \Leftarrow$ \DominantOrientation($O$).\;
$p^{best} \Leftarrow Null$; $s^{best} \Leftarrow -\infty$.\;
\For{$i=1$ to $K$}{
$h,s \Leftarrow$ \HypothesizePolicy($O,\theta$).\;
\lIf{$s>s^{best}$}{
$p^{best} \Leftarrow$ p; $s^{best} \Leftarrow s.$\;
}
}
}
\end{algorithm}
}

% \author{Nagasaka Tomomi and Tanaka Kanji% <-this % stops a space
\author{Kondo Kensuke ~~~~ Tanaka Kanji~~~~~~Nagasaka Tomomi% <-this % stops a space
\thanks{This work was partially supported by MECSST Grant in-Aid for
Young Scientists (B) (23700229), by KURATA grants and by TATEISI Science And Technology Foundation.}
\thanks{K. Kondo, K. Tanaka and T. Nagasaka are with Faculty of Engineering, University of Fukui, Japan.
        {\tt\small tnkknj@u-fukui.ac.jp}}%
	}

\CO{
\onecolumn
\textwidth 10cm
}

%\onecolumn

\maketitle
\thispagestyle{empty}
\pagestyle{empty}

%\onecolumn 
% \renewcommand{\FIG}[3]{~}\renewcommand{\FIGm}[3]{~}\renewcommand{\FIGR}[3]{~}\renewcommand{\FIGRm}[3]{~}\renewcommand{\FIGC}[5]{~}\renewcommand{\FIGf}[3]{~}\renewcommand{\FIGtpng}[5]{~}\renewcommand{\FIGpng}[5]{~}\renewcommand{\FIGRpng}[5]{~}

\newcommand{\hsa}{\hspace*{-2mm}}

\newcommand{\hsb}{\hspace*{-4mm}}

\newcommand{\myrule}[1]{\hspace*{-15mm} \mbox{[R#1] }}
\newcommand{\figgd}{
\begin{figure}[t]
\begin{center}
\FIG{6}{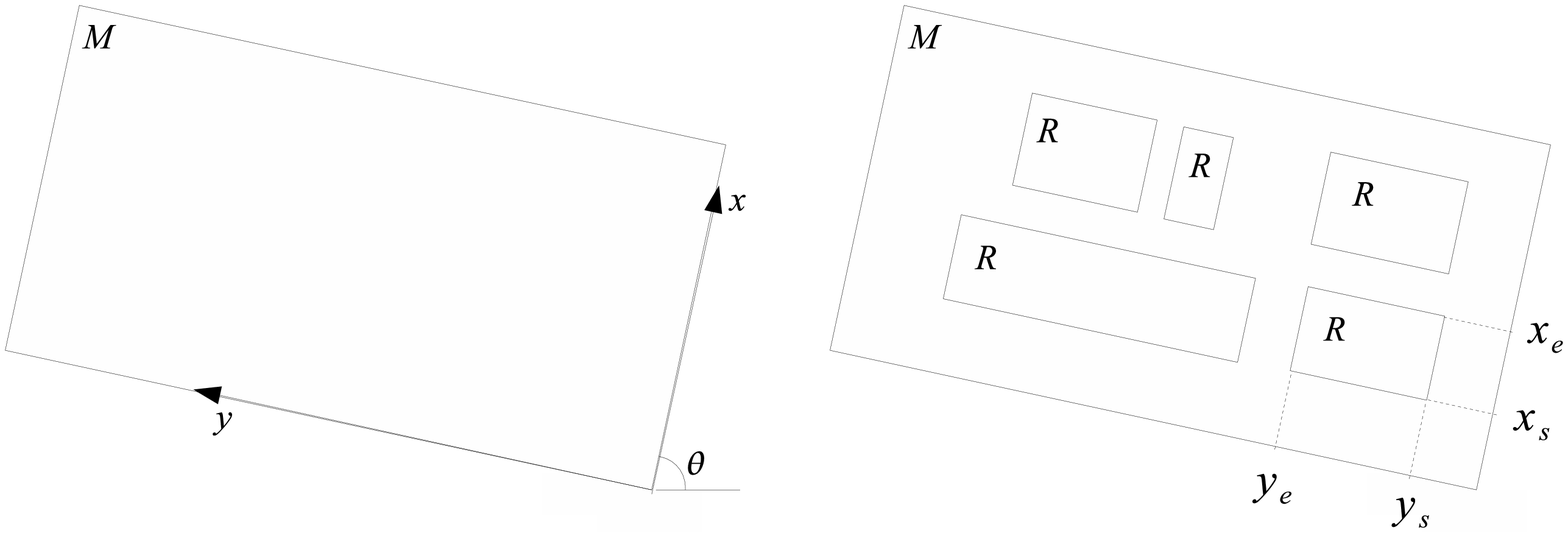}{[R1] \hspace*{3cm} [R2]}\vspace*{2mm}\\
\FIG{6}{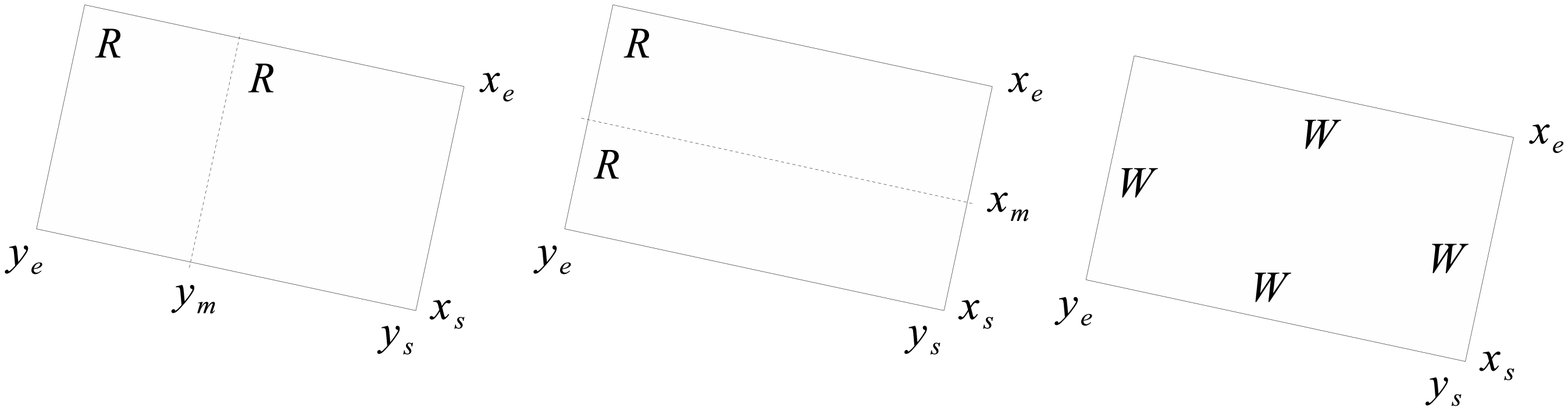}{\hspace*{2mm} [R3] \hspace*{1.2cm} [R4] \hspace*{1.2cm} [R5]}
{\scriptsize
\begin{eqnarray*}
&& \myrule{1} O \rightarrow M(\theta)^* \\
&& \myrule{2} M(\theta) \rightarrow R_{x_s,y_s,\theta}(x_s,y_s,x_e,y_e)^* \\
&& \myrule{3} R_{x,y,\theta}(x_s,y_s,x_e,y_e) \rightarrow \nonumber 
R_{x,y,\theta}(x_s,y_s,x_m,y_e) R_{x,y,\theta}(x_m,y_s,x_e,y_e) \\
&& \myrule{4} R_{x,y,\theta}(x_s,y_s,x_e,y_e) \rightarrow \nonumber 
R_{x,y,\theta}(x_s,y_s,x_e,y_m) R_{x,y,\theta}(x_s,y_m,x_e,y_e) \\
&& \myrule{5} R_{x,y,\theta}(a,b,c,d) \rightarrow \nonumber 
\bar{W}_{x,y,\theta,a,b,c,b} \bar{W}_{x,y,\theta,c,b,c,d} \bar{W}_{x,y,\theta,c,d,a,d} \bar{W}_{x,y,\theta,a,d,a,b} 
\end{eqnarray*}
}
\end{center}
\begin{center}
\begin{minipage}[b]{4cm}
\FIG{4}{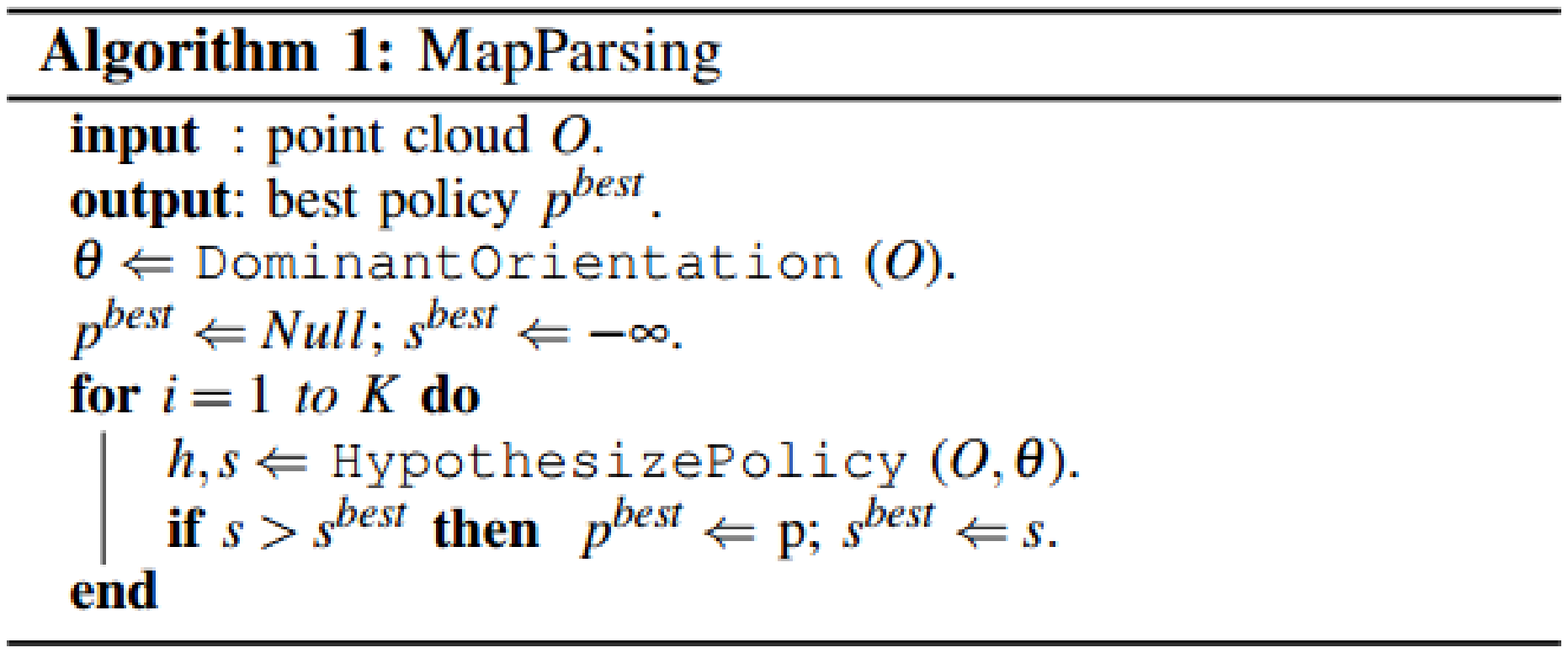}{}
\FIG{4}{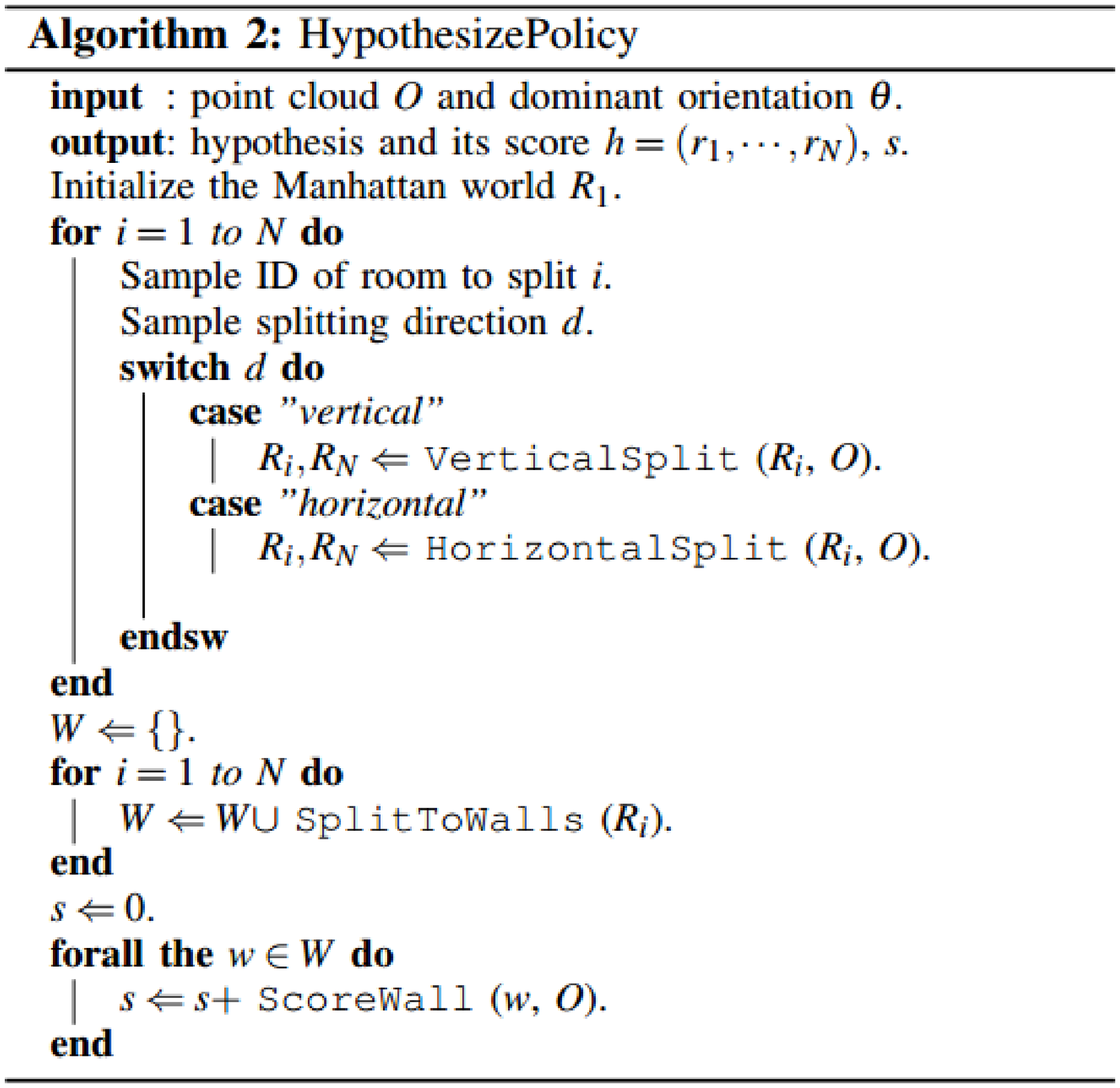}{}
\end{minipage}
\begin{minipage}[b]{4cm}
\FIG{4}{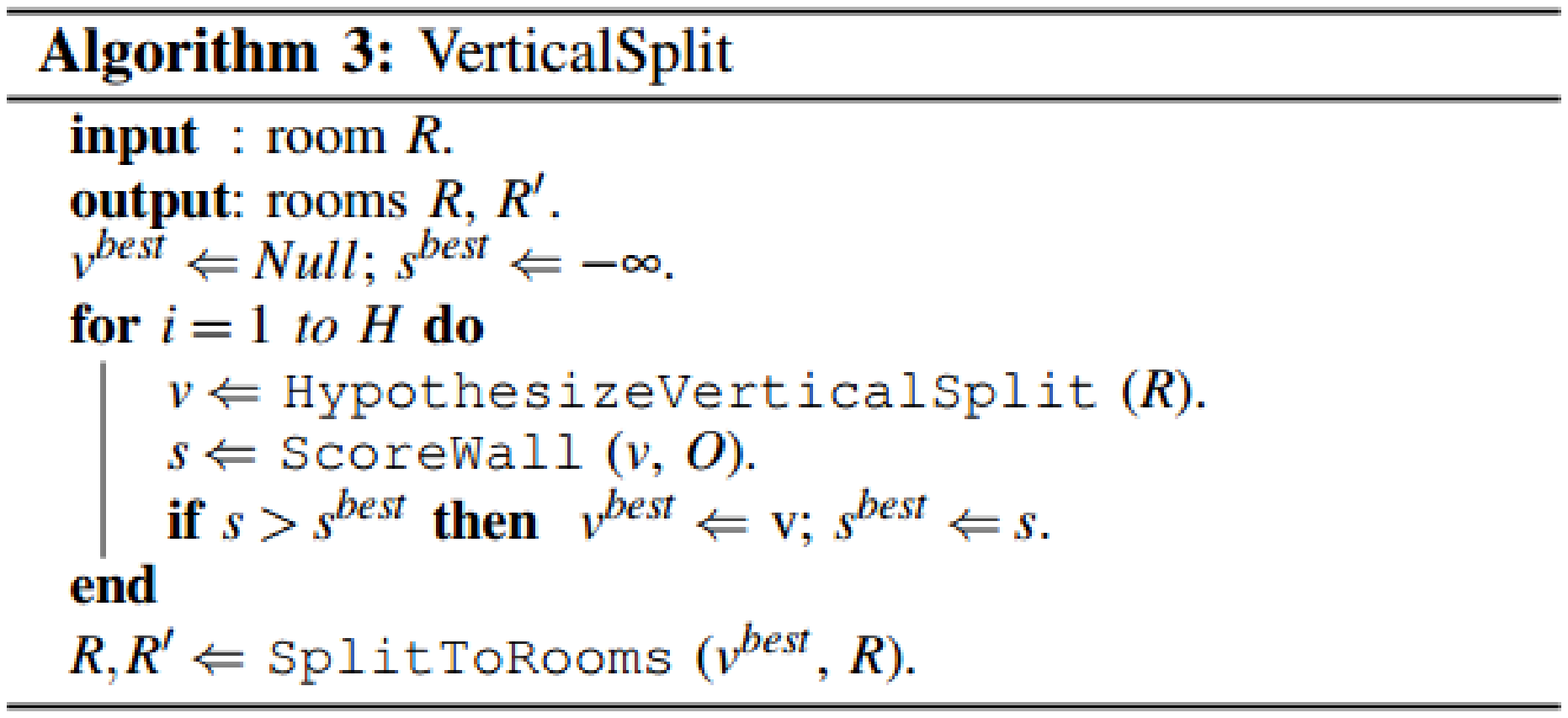}{}
\FIG{4}{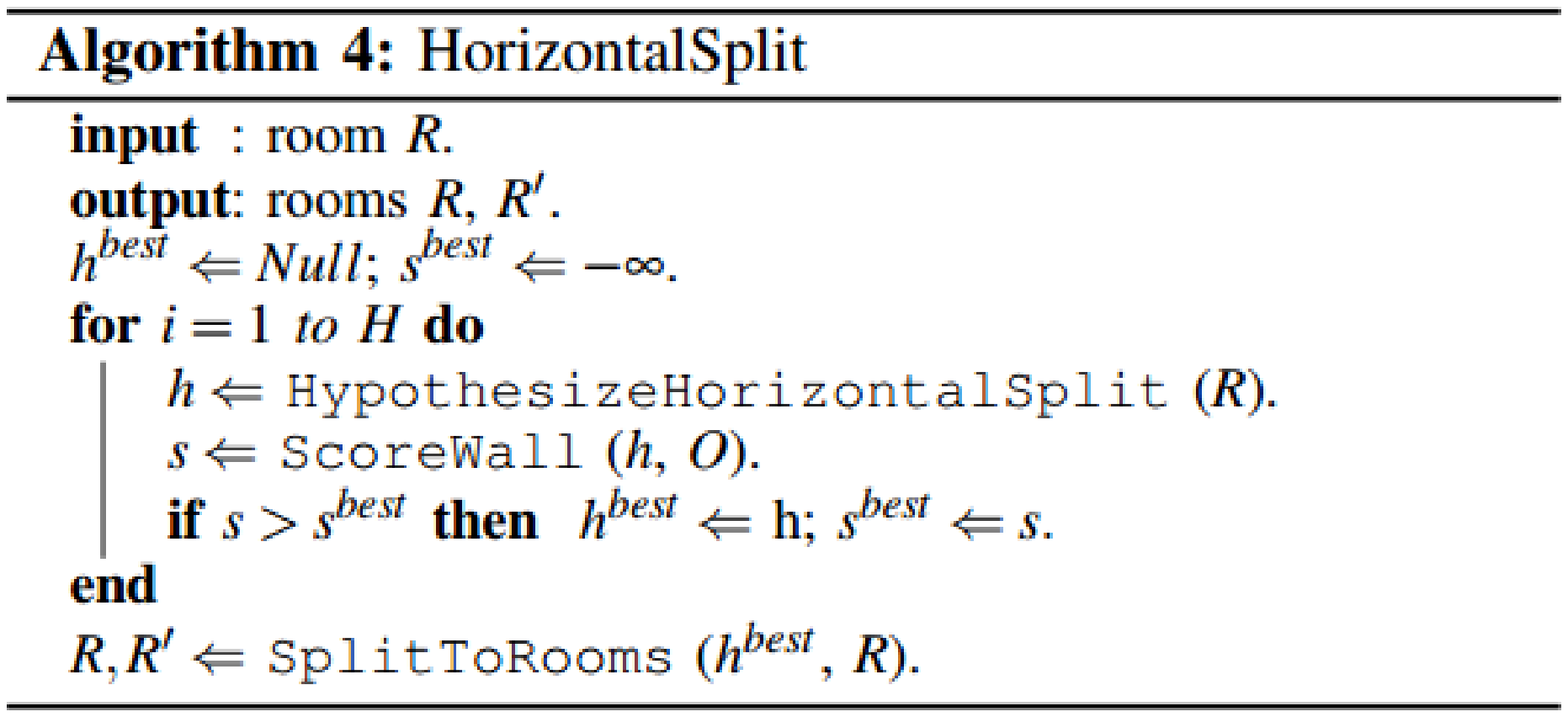}{}\\
\vspace*{1.9cm}
\end{minipage}\vspace*{-6mm}\\
%\\
%\FIGR{4}{code/figs/fr079_s_map.txt.eps}{fr079}
%\FIGR{4}{code/figs/kwing1_c_map.txt.eps}{kwing1}
\end{center}
\caption{Manhattan world grammar and map parsing. Top: A Manhattan world grammar
employing five rules, ``[R1] initialize a Manhattan world," ``[R2] split a Manhattan world into rooms," ``[R3] split a room vertically," ``[R4] split a room horizontally," and ``[R5] split a room into four orthogonal walls." Bottom: Map parsing algorithm. Examples of
map parsing are shown in Fig. \ref{fig:M}.}\label{fig:rules}\label{fig:snapshot}
\end{figure}
}

\newcommand{\hsf}{\hspace*{-4.2mm}}

\newcommand{\FIGpngj}[2]{\FIGpnga{2.4}{#1}{#2}}
\newcommand{\hsj}{\hspace*{-5mm}}

\newcommand{\figb}{
\begin{figure}[t]
\begin{center}
\FIGR{4}{fr079_s_map.txt-16.eps}{$N=16$}
\FIGR{4}{fr079_s_map.txt-128.eps}{$N=128$}
\FIGR{4}{fr079_s_map.txt-1024.eps}{$N=1024$}
\caption{Map parsing results for three different lengths of the rule sequence.}\label{fig:snapshot_n}
\end{center}
\end{figure}
}

\newcommand{\figca}{
\begin{figure}[t]
\begin{center}
\FIGR{7}{code/ratio_h.eps}{}
\vspace*{-5mm}
\caption{Compression ratio vs number $H$ of splitting hypotheses.}\label{fig:performance_a}
\end{center}
\vspace*{-5mm}
\end{figure}
}

\newcommand{\figcb}{
\begin{figure}[t]
\begin{center}
\FIGR{7}{code/ratio_n.eps}{}
\vspace*{-5mm}
\caption{Compression ratio vs length $N$ of the rule sequence.}\label{fig:performance_b}
\end{center}
\vspace*{-5mm}
\end{figure}
}

\newcommand{\figcc}{
\begin{figure}[t]
\begin{center}
\FIGR{7}{code/ratio_k.eps}{}
\vspace*{-5mm}
\caption{Compression ratio vs number $K$ of policy hypotheses generated.}\label{fig:performance_c}
\end{center}
\vspace*{-5mm}
\end{figure}
}

\newcommand{\figcd}{
\begin{figure}[t]
\begin{center}
\FIGR{6}{code/sec_n.eps}{}
\vspace*{-5mm}
\caption{Computation time vs length $N$ of the rule sequence.}\label{fig:performance_d}
\end{center}
\vspace*{-5mm}
\end{figure}
}

\newcommand{\figd}{
\begin{figure}[t]
\begin{center}
\scriptsize
\FIGR{6}{code/figs3/fr079_s_map.txt-0.200000.eps}{}\\
\FIGR{4}{code/ratio_n.0.2.eps}{}
\FIGR{4}{code/sec_n.0.2.eps}{}\\
res 0.2m\\
\FIGR{6}{code/figs3/fr079_s_map.txt-0.300000.eps}{}\\
\FIGR{4}{code/ratio_n.0.3.eps}{}
\FIGR{4}{code/sec_n.0.3.eps}{}\\
res 0.3m\\
\caption{Results for coarse resolutions.}\label{fig:coarse}
\end{center}
\vspace*{-5mm}
\end{figure}
}

\begin{abstract}
Map retrieval, the problem of similarity search over a large
collection of 2D pointset maps previously built by mobile robots,
is crucial for autonomous navigation in indoor and outdoor
environments. Bag-of-words (BoW) methods constitute
a popular approach to map retrieval; however, these methods have extremely limited
descriptive ability because they ignore the spatial
layout information of the local features. The main contribution of this paper is
an extension of the bag-of-words map retrieval method to enable the
use of spatial information from local features. Our strategy is to
explicitly model a unique viewpoint of an input local map; the pose of the local feature is defined with respect to this unique viewpoint, and can be viewed
as an additional invariant feature for discriminative map retrieval. Specifically, we wish to determine a unique viewpoint that is invariant
to moving objects, clutter, occlusions, and actual viewpoints.
Hence, we perform scene parsing to analyze the scene
structure, and consider the ``center" of
the scene structure to be the unique viewpoint. Our scene parsing is based on a Manhattan world
grammar that imposes a quasi-Manhattan world constraint to enable
the robust detection of a scene structure that is invariant to clutter
and moving objects. Experimental results using the publicly available
radish dataset validate the efficacy of the proposed approach.
\end{abstract}

\maketitle

\figN

\section{Introduction}\label{sec:1}

Map retrieval, the problem of similarity search over a large collection of local maps previously built by mobile robots, is crucial
for autonomous navigation in indoor and outdoor environments.
This study addresses a general map retrieval problem in which a 2D
pointset map is provided as a query, and the system searches a
size $N$ map database to determine similar database maps that are relevant
under rigid transformation. One of the most popular approaches to address
this problem is bag-of-words (BoW), a method derived from image retrieval
techniques \cite{ps1,nnl2,bow5}. In BoW, a collection of local invariant appearance
features (e.g., shape context \cite{sc1}, polestar \cite{polestar}) is extracted from an input map and each feature is translated into a visual word. Consequently, an input map is described compactly and matched efficiently as an
unordered collection of visual words, termed ``bag-of-words" \cite{ps1}.

A major limitation of the BoW scene model is the lack of spatial
information. The BoW methods ignore the spatial layout information of the features, and hence, they have severely limited
descriptive ability \cite{ps1}. Key relevant studies to address this issue include recent image retrieval techniques, such as spatial pyramid matching \cite{spm}. In such techniques, weak robust
constraints are extracted from the spatial information, and are incorporated into
the BoW model to significantly improve the discriminative power
of the model. However, to apply such methods that were originally
proposed for image data, we must first define the origin or
viewpoint of an input map with respect to which the poses of local features
are defined. This task is non-trivial because of the following reasons: (1) Unlike image data,
map data lacks an explicit viewpoint; (2) Unlike image data, the area
of a map is variable; it is incrementally updated by mapper robots and
can grow in an unbounded manner.

The main contribution of this study is an extension of the BoW map retrieval method to enable the use of spatial information from
local features (Fig. \ref{fig:N}). Our strategy is to explicitly model a unique
viewpoint of an input local map; the pose of the local
feature is defined with respect to this unique viewpoint, and can be viewed as an additional invariant feature
for discriminative map retrieval. Specifically, we wish to determine
a unique viewpoint that is invariant to moving objects, clutter,
occlusions, and actual viewpoints. Hence, we perform scene parsing
to analyze the scene structure, and consider the ``center" of the scene structure to be the unique viewpoint. Our scene parsing is based on
a Manhattan world grammar that imposes a quasi-Manhattan world
constraint to enable the robust detection of a scene structure that is invariant
to clutter and moving objects. We also discuss several strategies
for extracting the ``center" of a given scene structure. We generated
a database of 2D local maps built by mobile robots from the publicly
available radish dataset \cite{radish}, and experimentally validated the efficacy
of the proposed method.

\subsection{Related Work}

Existing approaches to scene retrieval can be classified according to
the feature descriptors used, the manner in which the feature descriptors are used, and whether the
feature approach is global or local. A global feature approach describes
the global structure of a scene by using a single global feature descriptor
(e.g., Gist, HOG). In contrast, a local feature approach describes a
scene by using a collection of local feature descriptors (e.g., SIFT). In
general, both the approaches can be used complementarily; however, the
focus of this paper is on the local feature approach.

Direct feature matching \cite{neira03, tipaldi2010flirt, m3rsm}
and BoW \cite{ps1,nnl2,bow5} are two popular local feature approaches. 
In \cite{neira03},
the authors introduce the concept of RANSAC map matching
for loop closure detection using pointset maps.
In \cite{tipaldi2010flirt}, the authors detect interest
points, extract appearance descriptors (e.g., shape context), and perform a direct match
between the appearance descriptors of the query and database images. 
Very recently,
in \cite{m3rsm}
the authors presented
an efficient direct matching 
based on multi-resolution many-to-many map matching framework.
In \cite{kanji09}, we also
addressed the scalability issue by introducing a pre-filter based on
the appearance descriptor. 
However, the direct matching methods have limited scalability because they require a large amount
of time and space that is linearly proportional 
to the number of maps.

BoW methods \cite{ps1,nnl2,bow5} are
well known for efficient map retrieval. In these methods, an
input map is described compactly by an unordered collection of vector
quantized appearance descriptors. 
In \cite{iros15a},
we alse employed a bag-of-words scene model to achieve efficient visual robot localization.
However, their descriptive ability is
limited because they ignore all the layout information of the local
features. We address this limitation in our study.

A majority of the existing BoW map retrieval methods 
explicitly or implicitly assume that the viewpoint trajectory of the
mapper robot with respect to the local map is unavailable \cite{ps1}. In contrast, we explicitly use the viewpoint information produced by our viewpoint
planner as a cue to compute the local map descriptor. The success
of our approach is based on the assumption that the viewpoint planner
provides a unique viewpoint for a local map; therefore, we also
consider viewpoint planning. 
These two issues have not been explored in existing literature.

Our study is also similar to several 
image retrieval techniques that
describe the appearance and spatial information of local
features. Among these methods, the part model \cite{part2008}, in which a scene
is modeled as a collection of visual parts, is extremely popular. Spatial
pyramid matching is an alternative method that places a sequence of increasingly coarser
grids over the image region, and considers a weighted sum of the number
of matches that occur at each level of resolution. However, most existing studies focus on image data, and
 do not handle map data that has no explicit viewpoint, as we
discussed in Section \ref{sec:1}.

Our map parsing method can be viewed as an instance of scene parsing,
which has been extensively studied in the fields of point-based geometry
\cite{octree2006}, image description \cite{i2t}, scene reconstruction \cite{instantarchitecture}, and
scene compression \cite{scenecompression}. Scene parsing approaches are broadly classified as generic
approaches (e.g., line primitives \cite{lineprimitives2002}, plane primitives \cite{planeprimitives2004}, etc.) and
parametric approaches (e.g., constructive solid geometry \cite{csg2007}, hierarchical model \cite{hierarchical2003}, grammar-based \cite{shapegrammar2006}, etc.). Our study can be viewed
as a novel application of scene parsing to the map retrieval problem.

This study is a part of our studies on loop closure detection \cite{kanji06} and map-matching \cite{shogo2013partslam}, and related to our previous works in ICRA15, IROS15, and PPNIV15 papers \cite{iros15a,icra15a, ppniv15}. However, the use of viewpoint planning in map retrieval tasks is not addressed in existing studies.

\section{Map Retrieval Approach}\label{sec:baseline}

For clarity of presentation, we first describe the overview of
the map retrieval
system that is the basis for our approach and is a
performance comparison benchmark in the experiments described in
Section \ref{sec:exp}. The main steps in the process 
are as follows: (1) Building informative local maps, (2) Planning
the unique viewpoint of the local map, (3) Constructing a 
local map descriptor (LMD), and (4) Indexing/Retrieving the map database from
the LMD descriptors. These four steps are explained below.

\subsection{Map Building}\label{sec:2a}

Based on existing literature \cite{paz2008large}, 
we build a local map from a short sequence of perceptual and odometry measurements; each measurement sequence must be sufficiently long to capture the rich appearance and geometric information of the local surroundings of the robot. In
the implementation, each sequence corresponds to a 5 m run of the robot. Any map-building algorithm (e.g., FastSLAM, scan matching) can be used to register a measurement sequence to a local map. We start generating a local map every time the viewpoint of the robot moves 1 m along the path. Thus, a collection of overlapping local maps along the path is generated.

\subsection{Viewpoint Planning}\label{sec:2b}

In order to determine a unique viewpoint of a given input map that is
invariant to moving objects, clutter, occlusions, and actual viewpoints, we first perform scene parsing to analyze the scene structure. Then, we consider the ``center" of the scene structure to be the unique viewpoint. We
will discuss several strategies for viewpoint planning in Section \ref{sec:3}.
For example, in the strategy $S^1$, the scene structure is first analyzed to obtain a set of points, termed ``structure points," that belong to a structure (e.g., walls); then, the center-of-gravity of the structure points is computed, and
finally, a nearest-neighbor unoccupied location relative to the center of gravity is determined as the
unique viewpoint.

\subsection{Map Description}\label{sec:2c}

We follow a standard BoW approach \cite{ps1}
for extracting and representing
appearance features. We adopt the polestar feature
because it has several desirable properties, including viewpoint invariance and rotation independence, and has proven effective as a
landmark for map matching in previous studies \cite{tanaka2012multi}. The extraction algorithm consists of three steps: (1) First, a set of keypoints is
sampled from the raw 2D scan points. (2) Next, a circular grid is
imposed and centered at each keypoint with different $D=10$ radius.
(3) Finally, the points located in each circular grid cell are counted,
and the resulting $D$-dim vector is generated as the output, the polestar descriptor.
We quantize the appearance descriptor (i.e., $D$-dim polestar vector) of
each feature to a 1-dimensional code termed an ``appearance word". This
quantization process consists of three steps: (1) normalization of the
$D$-dim vector by the L1 norm of the vector, (2) binarization of each $i$-th
element of the normalized vector into $b_i\in \{0,1\}$, and (3) translation of
the binarized $D$-dim vector into a code or a visual word, $w_a=\sum_i 2^i b_i$.
Currently, the threshold for binarization is determined by calculating the mean of all the elements of the vector. Thus, a map is represented
by an unordered collection of visual words,
$\{w_a ~ | ~ w_a \in [1,K] \}$,
called BoW.
We consider
$D$-dim 
binarized polestar descriptors, and hence, the
vocabulary size is $K=2^{10}$.

In order to translate the pose of each feature with respect to the planned unique
viewpoint to a visual word, we quantize the pose or keypoint of the
feature by using a resolution quantization step size of 0.1 m to obtain a
code, $(w_x, w_y)$, termed ``pose word".

Finally, we obtain a BoW representation of the input local
map, termed ``local map descriptor (LMD)". An LMD is an unordered
collection of visual words, each having the form:
\begin{equation}
\langle w_x,w_y,w_a \rangle \label{eqn:proposed}.
\end{equation}

\subsection{Map Indexing/Retrieval}\label{sec:2d}

In order to index and retrieve the BoW map descriptors, we use the appearance word $w_a$ as the primary index for the inverted file system, and the pose word 
$(w_x, w_y)$ as an additional cue for fine matching. 
The retrieval stage begins with a search of the map collection. The given appearance word
$w_a$
is used as a query to obtain all the memorized
feature points with common appearance words, and to filter out the
feature points whose pose word
$(w_x',w_y')$
is distant from that of the query feature
$(w_x,w_y)$:
\begin{equation}
| w_x - w_x' | > D_{x,y}, 
\end{equation}
\begin{equation}
| w_y - w_y' | > D_{x,y}.
\end{equation}
Thus, the final shortlist of maps is obtained. Currently, we use a large threshold,
$D_{x,y}=3[m]$,
to suppress false negatives, i.e., incorrect identification of
relevant maps as not being relevant.

We use the BoW representation for the database construction
and retrieval processes. In the database construction process, each local map is
indexed by the inverted file system; each word $w_a$
belonging
to the map is used as an index. In the retrieval process, 
all the indexes that
have words in common with the query map are accessed, and the
resulting candidate database maps are ranked based on the frequency
or the number of words matched. 
For 
$K$ words in the vocabulary,
a frequency histogram of visual words is represented by a 
$K$-dim vector.

\figgd

\section{Viewpoint Planning}\label{sec:3}

In order to determine a unique viewpoint that is invariant to
moving objects, clutter, and actual viewpoints, we first parse the scene
structure using a Manhattan world grammar (Fig. \ref{fig:rules}), 
and then, determine the
unique viewpoint with respect to the structure points. In the following subsections,
first, we briefly introduce the Manhattan world grammar. Then, we
describe the scene parsing algorithm, and discuss several strategies
for viewpoint planning.

\subsection{Manhattan World Grammar}\label{sec:3a}

We use the formulation of context free grammar (CFG) to
implement the Manhattan world grammar. CFG defines the grammar
as
\begin{equation}
G=(V, T, R, U),
\end{equation}
where $V$ is a set of non-terminal nodes. Each non-terminal node is
represented by a capital letter, 
etc. 
$T$ is a set of
terminal nodes. Each terminal node is represented by a capital letter
with a bar over it,
e.g., '$\bar{A}$', '$\bar{B}$', '$\bar{C}$', etc.

In the case of a map parsing problem,
either a terminal or a non-terminal node is modeled as a geometric
primitive; in our study, we use a ``room" or ``wall" primitive. $R$ is a set of
replacement rules. Each replacement rule
$r\in R$ is in the form
\begin{equation}
A \rightarrow a
\end{equation}
and replaces a non-terminal node $A$ with a sequence of terminal or non-
terminal nodes $a$. $U$ is the start variable. Let constant $N$ denote the upper
bound on the number of grammar rules applied in a map parsing task.
Let variable $r_i$ denote the $i$-th rule in the length $N$ rule sequence. The
solution space of a map parsing problem is defined as
\begin{equation}
p = \{(r_1, r_2, \cdots, r_N)\}.
\end{equation}
The score $S(p)$ $(p\in P)$
of a policy $p$ is evaluated in terms of 
how
well the original input map is explained by a set of primitives (i.e.,
terminal nodes) produced by the rule sequence $p$. 
The objective of a
map parsing problem is to find a ``best" policy
$p^{best}(\in P)$
that maximizes
the score function $S(p)$. In our method, the score
$S(p)$
is evaluated in terms of the ratio of datapoints that are explained by the
rule sequence $p$:
$Size(O\setminus O^c)/Size(O)$,
where
$Size(\cdot)$ 
is the number of
datapoints, $O$ is the set of datapoints in the input map, and
$O^c(\subset O)$
is a subset that is explained by the grammar $p$.

In order to adapt CFG to our map parsing method, we model the entire map as a set
of ``Manhattan worlds," and ``rooms" and ``walls". A Manhattan
world \cite{MWorg} is a set of rectangular rooms aligned with the orthogonal
directions. A room is composed of a set of four orthogonal walls. A
wall is represented by a 2D line segment. The grammar is represented
by a set of rules, R1, R2, R3, R4, and R5. Figure \ref{fig:rules} illustrates each
rule and its parameter settings. The symbol $O$ represents the
original pointset map. A Manhattan world
$M(\theta)$
is explained by a
collection of orthogonal room primitives and is oriented at angle
$\theta$.
A room primitive
$R_{\theta}(x,y,w,h)$
is explained by smaller orthogonal
rooms or a set of four orthogonal walls. The angle, width, and
height of a room primitive are
$\theta$, $w$ and $h$, respectively. A wall
primitive
$\bar{W}_{x,y,\theta,a,b,c,d}$
is represented by a straight line segment, which
is the result of rotating a line segment
$(a,b)$-$(c,d)$
by angle $\theta$ around
the point $(x,y)$.

\figM

\subsection{Map Parsing}

Our method for determining a best policy uses a hypothesize-and-verify approach. Given an input pointset map, we wish to determine a policy  that maximizes a pre-defined score function $p^{best}(\in P)$.
Our approach first estimates the dominant orientation $\theta$ of the Manhattan
world; then, it generates $K$ random hypotheses for the policy, assigns a score to each
hypothesis, and selects the hypothesis with the highest score as the best policy.
The above methods for estimating the dominant orientation, generating
hypotheses, and assigning a score to the hypotheses are explained in algorithms I-IV in Fig. \ref{fig:rules}. 
As mentioned, the score $S(\cdot)$ of a policy hypothesis is defined by the number of datapoints that would be explained by the policy. 
A data-driven algorithm
is used for policy generation and evaluation.

\subsection{Viewpoint Planning}

Given a scene understanding from the grammar-based parsing,
we identify the ``center" of a map and use it as the unique viewpoint
(UVP), as shown in Fig. \ref{fig:M}. In this study, we implement five 
strategies, $S^1$-$S^5$, to identify the center of a given input map, and
experimentally evaluate the effectiveness of each strategy in terms of
``viewpoint uniqueness" and map retrieval performance.

In this subsection, we use the following technical terms: grid map, free cells,
unknown cells, wall cells, structure cells, and unoccupied cells. A grid
map is a classical representation of a map that imposes a discretized
grid on the $xy$-plane and classifies each cell as occupied, free, or
unknown \cite{thrun2005probabilistic}. We denote occupied, free, and unknown cells as 
$C^{occupied}$, $C^{free}$ and $C^{unknown}$, respectively.
The grid map is constructed during the map building process described in 
Section \ref{sec:2a}.
In addition,
wall, structure, and unoccupied cells are defined based on the 
three cell classes mentioned above.
Wall cells, $C^{wall}$, are the cells that are occupied by
the wall primitives defined in Section \ref{sec:3a}. Structure cells, $C^{structure}$, are defined as
$C^{structure}=C^{occupied} \cap C^{wall}$. 
Unoccupied cells, $C^{unoccupied}$
are defined as 
$C^{unoccupied}=C^{free} \setminus C^{structure}$.

The  strategy $S^1$ determines UVP as an unoccupied location near the center-of-gravity of structure points. 
First, 
$S^1$ parses the scene structure and obtains a set of wall points;
then, 
it computes structure points 
$C^{structure}=\{(x_i, y_i)\}_{i=1}^N$ 
from the wall points and occupied cells as
shown above, and, based on the result, it computes the center-of-gravity
of the structure points
$p^{cog}=N^{-1}\sum_{i=1}^N[x_i~y_i]^T$. 
Finally, it searches the
unoccupied cells and determines a nearest-neighbor unoccupied cell
relative to the center-of-gravity to be UVP:
$p^{uvp}=\arg \min_{v\in C^{unoccupied}}|v-p^{cog}|$.

The strategy $S^2$ determines UVP as an unoccupied cell that minimizes the distance to the farthest structure points. Similar to the strategy $S^1$, $S^2$ parses the scene structure to obtain the structure and unoccupied cells. Then, for each viewpoint candidate $v$ (i.e., unoccupied cell), $S^2$ evaluates the distance
$d(v)=\max_{p\in C^{structure}} |v-p|$
between the viewpoint and
its farthest structure point, and selects one candidate that minimizes
the evaluated distance:
$p^{uvp}=\arg \min_{v\in C^{unoccupied}}d(v)$.

The strategy $S^3$ determines UVP as an unoccupied cell that maximizes the distance to the nearest structure points. The process for viewpoint planning is similar to that in $S^2$. The only difference is that $S^3$ uses the minimum distance instead of the maximum distance, and the maximum operator instead of the minimum operator, i.e., 
$d(v)=\min_{p\in C^{structure}} |v-p|$,
and 
$p^{uvp}=\arg \max_{v\in C^{unoccupied}}d(v)$.

The strategy $S^4$ is based on the analysis of dominant structures, which
are defined as the longest line segments on the input map. This strategy
is similar to $S^1$; however, instead of using every structure point, $S^4$ uses 
only the 10 longest walls to compute the structure cells.

The strategy $S^5$ determines UVP as the center of the unoccupied regions.
In the viewpoint planning process, $S^5$ searches a set of bounding boxes of unoccupied cells aligned with the orthogonal directions of the Manhattan world; then, it generates two histograms $f^x,f^y$ 
of unoccupied cells along two dominant directions of the Manhattan world. Next, the peaks
$x^*=\arg \max_x f^x(x)$ and $y^*=\arg \max_y f^y(y)$ 
of the two histograms are searched. Further, a bounding box of unoccupied cells $(x,y)$ whose $f^x(x)$ and $f^y(y)$ values exceed $0.9 f^x(x^*)$ and $0.9 f^y(y^*)$, respectively,
is computed.
Finally, UVP is defined as the center of the bounding box.

\figD
\figO

\section{Experiments}\label{sec:exp}

We conducted map retrieval experiments to verify the efficacy of
the proposed approach. In the following subsections, first, we describe
the datasets and the map retrieval tasks used in the experiments; then, we
present the results and compare the performance of our method with that of other methods.

\subsection{Dataset}

For map retrieval, we created a large-scale map collection from
the publicly available radish dataset \cite{radish}, which comprises odometry
and laser data logs acquired by a car-like mobile robot in indoor
environments (Fig. \ref{fig:D}). We used a scan matching algorithm to create a collection of query-database maps
from each of six datasets
---``albert," ``fr079," ``run1," ``fr101," ``claxton," and ``kwing"---
that were obtained from 
212, 209, 80, 277, 79, and 286 m travel of the mobile robot,
corresponding to 
4167, 3118, 2882, 5299, 4150, and 609 scans. 
Fig. \ref{fig:O}
% Figs. \ref{fig:Bbca} and \ref{fig:Bbcb} 
shows examples of the query and database maps. The map collection comprises more than
1065 maps. Our map collections contain many virtually duplicate
maps, thus making map retrieval a challenging task.
We use ``claxton" only as additional distructer maps 
for increasing the database size,
as ``claxton" does not contain any loop closure.

\subsection{Qualitative Results}

The objective of map retrieval is to find a relevant map
from the map database for a local map given as a query. The relevant
map is defined as a database map that satisfies two conditions: (1)
Overlap of datapoints between the query and the relevant maps exceeds
$R^{overlap}=75\%$, 
and (2) Its distance traveled along the robot's trajectory is distant from that of the query map, such as in a ``loop-closing" situation
in which a robot traverses a loop-like trajectory and returns to a
previously explored location.

For each relevant map pair, map retrieval is performed using
a query map and a size $N$ map database, which consists of
the relevant map and $(N-1)$ random irrelevant maps. The spatial
resolution of the occupancy map is set to 0.1 m. We implemented the
map retrieval algorithm in C++, and successfully tested it with various
maps.
Fig. \ref{fig:O} shows the results of map retrieval performed using the baseline
(``BoW") and the proposed (``LMD") systems. 
As described in Section \ref{sec:baseline},
BoW differs from LMD only in that
it does not use pose word
but only uses appearance word.
It is observed that the proposed LMD method yields fewer
false positives than
the BoW method. The reason for this result is that, in the proposed LMD method, many incorrect matches are
successfully filtered out by the proposed descriptor, which uses the
keypoint configuration as a cue. 
It can be observed that, for these examples, the proposed LMD method using the spatial layout of
local features as a cue is successful in finding relevant maps.

\tabAd

\figR

\subsection{Quantitative Results}

For performance comparison, we evaluated the averaged normalized
rank (ANR) \cite{shogo2013partslam} for the BoW and LMD methods. ANR is
a ranking-based performance measure in which a lower value is
better. In order to determine the ANR, we performed several independent
map retrieval tasks with various queries and databases. For each
task, the rank assigned to the ground-truth database map by a map
retrieval method of interest was investigated, and the rank was normalized by the
database size $N$. The ANR was subsequently obtained as the average of
the normalized ranks over all the map retrieval tasks. All map retrieval
tasks were conducted using 247 different queries and a size 1065 map
database.

Table \ref{tab:A} and Fig. \ref{fig:R} summarize the ANR performance. The proposed LMD
system with strategies $S^1$, $S^2$, $S^4$, and $S^5$ 
clearly 
outperforms the baseline BoW system. 
An exception is the strategy $S^3$
and will be discussed in the next subsection, Section \ref{sec:st_exp}.
By filtering
out incorrect matches using the keypoint configuration as a cue, the
LMD method was able to successfully perform map retrieval in many
cases, as shown in the table. In contrast, the BoW system based only on
appearance words does not perform well in many cases, mainly owing to the large number of false matches. The above results verify the
efficacy of our approach.

\CO{

\subsection{Parameter Sensitivity}\label{sec:sensitivity}

\tabC

Table \ref{tab:C}
shows results for different settings
of the threshold $D_{x,y}$
on pose word.
One can see that
performance of strategy $S^1$ becomes low
when too small threshold $D_{x,y}=1[m]$ is used,
and performance of strategy $S^5$ is more stable.
A main reason is that 
because 
the strategy $S^5$ is based on analysis of 
both structure and unoccupied regions,
the UVP is often invariant against actual viewpoints.

}

\subsection{Comparing Different Strategies}\label{sec:st_exp}

Table \ref{tab:A}
also compares different viewpoint planning strategies for 
the proposed LMD algorithm.
One can see that
$S^4$ and $S^5$ are best strategies in the current experiment.
The strategy $S^4$ is based on dominant structure in the map
and it was successful in finding center of structures.
The strategy $S^5$ is based on analysis of 
unoccupied regions and it was often successful in finding
center of unoccupied regions.
On the other hand,
$S^3$ was not as good as other strategies 
and the BoW method.
A main reason is that 
because 
$S^3$
maximizes the distance 
from UVP to the nearest structure points,
it often determines UVP 
near the boundary between
free and unknown region,
which is naturally far apart from 
the center of the map.
On the other hand,
$S^2$
provided a good result
as it minimizes the distance 
from UVP to the farthest structure points,
which is often located at the center of a map.
Finally,
$S^1$
uses all the datapoints in a map
and tends to be influenced by non-structure points and noises,
and as a result,
it performs not as good as
$S^4$.

\figQ

\subsection{Viewpoint Planning}

In this subsection, we investigate the performance of our viewpoint
planning method. As mentioned earlier, the success of our approach is
based on the assumption that the viewpoint planner provides a unique
viewpoint for a given local map. As a proof-of-concept experiment, we
investigate the similarity between the planned viewpoints of the query and
those of the relevant database maps. We performed viewpoint planning
for the 247 pairs of query and relevant maps, and computed the errors
in the viewpoints planned. Fig. \ref{fig:Q} shows a summary of the investigation in the form of a histogram. The difference between the planned viewpoint of the query and that of the relevant database maps was,
for 90\% of the viewpoints considered in the current study,
within 5 m, 7.8 m, 12 m, 6.2 m, and 6.4 m for strategies $S^1$-$S^5$, respectively.

\subsection{Matching Visual Words}

\figBca
\figBcb

Figs. \ref{fig:Bbca} and \ref{fig:Bbcb} show 
the results of matching visual words using the baseline (``BoW") and the proposed (``LMD") systems. %
In these figures,
purple and green points indicate the query and the database maps, while the red lines indicate correspondence found by either method. To facilitate visualization, both maps are aligned w.r.t. the true viewpoints.
With the above visualization,
one can recognize false positive matches 
produced by
either BoW or LMD method
as they appear as relatively long red line segments
that connect wrong pairs of 
datapoints between query and database maps.
One can see that
LMD methods provide
significantly less amount of matches 
for irrelevant pairs
than for relevant pairs
comparing to BoW method.

\tabB

\subsection{Dissimilar Map Pairs}

As a final investigation,
we conducted an additional experiments 
on a challenging map retrieval scenario.
In this study,
we are interested in 
how 
robust
individual map retrieval algorithms are
and how well 
they perform 
on retrieving dissimilar maps.
To this end,
we use a lower threshold of overlap 
$R^{overlap}=50\%$,
instead of the previous setting $R^{overlap}=75\%$.
Table \ref{tab:B}
reports the ANR performance.
The strategies
$S^4$ and $S^5$
again
performed well
and 
$S^1$ was best performed in this case.
We can observe that
despite the challenging setting,
the proposed algorithm
is still successful in viewpoint planning
and map retrieval.

\section{Conclusions}

In this study, we focused on a method that extends BoW map
retrieval to enable the use of spatial information from local features.
Our strategy is to explicitly model a unique viewpoint of an input local
map; the pose of the local feature is defined with respect to this unique viewpoint, and can be
viewed as an additional invariant feature for discriminative map retrieval.
Specifically, we wish to determine a unique viewpoint that is invariant
to moving objects, clutter, occlusions, and actual viewpoints. Hence, our approach employs scene parsing to analyze the scene
structure, and the ``center" of
the scene structure is determined to be the unique viewpoint. Our scene parsing method is based on a Manhattan world
grammar that imposes a quasi-Manhattan world constraint to enable
the robust detection of a scene structure that is invariant to clutter
and moving objects. We have also discussed several strategies for
viewpoint planning that are based on different definitions of the ``center" of a map.
Experimental results using the publicly available radish dataset validate
the efficacy of the proposed approach.

\bibliographystyle{unsrt}
\bibliography{gmc}

\begin{thebibliography}{10}

\bibitem{ps1}
M.~Cummins and P.~Newman.
\newblock Appearance-only slam at large scale with fab-map 2.0.
\newblock {\em Int. J. Robotics Research}, 30(9):1100--1123, 2011.

\bibitem{nnl2}
Jan Knopp, Josef Sivic, and Tomas Pajdla.
\newblock Avoiding confusing features in place recognition.
\newblock In {\em ECCV}, pages 748--761. 2010.

\bibitem{bow5}
Panu Turcot and David~G Lowe.
\newblock Better matching with fewer features: The selection of useful features
  in large database recognition problems.
\newblock In {\em Computer Vision Workshops (ICCV Workshops), 2009 IEEE 12th
  International Conference on}, pages 2109--2116. IEEE, 2009.

\bibitem{sc1}
Mori G., Belongie S., and Malik J.
\newblock Shape contexts enable efficient retrieval of similar shapes.
\newblock {\em Proc. IEEE Computer Vision and Pattern Recognition}, pages
  723--730, 2001.

\bibitem{polestar}
E.~Silani and M.~Lovera.
\newblock Star identification algorithms: Novel approach \& comparison study.
\newblock {\em IEEE Trans. Aerospace and Electronic Systems}, 42(4):1275--1288,
  2006.

\bibitem{spm}
Svetlana Lazebnik, Cordelia Schmid, Jean Ponce, et~al.
\newblock Spatial pyramid matching.
\newblock {\em Object Categorization: Computer and Human Vision Perspectives},
  3:4, 2009.

\bibitem{radish}
Andrew Howard and Nicholas Roy.
\newblock The robotics data set repository (radish), 2003.

\bibitem{neira03}
Neira J., Tardos J.D., and Castellanos J.A.
\newblock Linear time vehicle relocation in slam.
\newblock {\em Proc. IEEE Int. Conf. Robotics and Automation}, 1:427-- 433,
  2003.

\bibitem{tipaldi2010flirt}
Gian~Diego Tipaldi and Kai~O Arras.
\newblock Flirt-interest regions for 2d range data.
\newblock In {\em Robotics and Automation (ICRA), 2010 IEEE International
  Conference on}, pages 3616--3622. IEEE, 2010.

\bibitem{m3rsm}
Edwin Olson.
\newblock {M3RSM:} many-to-many multi-resolution scan matching.
\newblock In {\em {IEEE} International Conference on Robotics and Automation,
  {ICRA} 2015, Seattle, WA, USA, 26-30 May, 2015}, pages 5815--5821, 2015.

\bibitem{kanji09}
K.~Saeki, K.~Tanaka, and T.~Ueda.
\newblock Lsh-ransac: An incremental scheme for scalable localization.
\newblock {\em Proc. IEEE Int. Conf. Robotics and Automation}, pages
  3523--3530, 2009.

\bibitem{iros15a}
Tanaka Kanji.
\newblock Cross-season place recognition using nbnn scene descriptor.
\newblock In {\em Intelligent Robots and Systems, 2015 IEEE/RSJ International
  Conference on}. IEEE, 2015.

\bibitem{part2008}
Dongfeng Han, Wenhui Li, and Zongcheng Li.
\newblock Semantic image classification using statistical local spatial
  relations model.
\newblock {\em Multimedia Tools Appl.}, 39(2):169--188, 2008.

\bibitem{octree2006}
Ruwen Schnabel and Reinhard Klein.
\newblock Octree-based point-cloud compression, 2006.

\bibitem{i2t}
B.Z. Yao, X.~Yang, L.~Lin, M.W. Lee, and S.C. Zhu.
\newblock I2t: Image parsing to text description.
\newblock 98(8):1485--1508, 2010.

\bibitem{instantarchitecture}
Peter Wonka, Michael Wimmer, Fran{\c c}ois Sillion, and William Ribarsky.
\newblock Instant architecture.
\newblock {\em ACM Transaction on Graphics}, 22(3):669--677, 2003.

\bibitem{scenecompression}
Alexander Toshev, Philippos Mordohai, and Ben Taskar.
\newblock Detecting and parsing architecture at city scale from range data.
\newblock {\em Computer Vision and Pattern Recognition, IEEE Computer Society
  Conference on}, 0:398--405, 2010.

\bibitem{lineprimitives2002}
Stephan Scholze, Theo Moons, and Luc J.~Van Gool.
\newblock A probabilistic approach to building roof reconstruction using
  semantic labelling.
\newblock {\em Proc. the 24th DAGM Symposium on Pattern Recognition}, pages
  257--264, 2002.

\bibitem{planeprimitives2004}
Zuwhan Kim and Ramakant Nevatia.
\newblock Automatic description of complex buildings from multiple images.
\newblock {\em Comput. Vis. Image Underst}, 96:2004, 2004.

\bibitem{csg2007}
Mathieu Br{\'e}dif, Didier Boldo, Marc~Pierrot Deseilligny, and Henri
  Ma\^{\i}tre.
\newblock 3d building reconstruction with parametric roof superstructures,
  2007.

\bibitem{hierarchical2003}
Sung~Chun Lee and Ram. Nevatia.
\newblock Interactive 3d building modeling using a hierarchical representation.
\newblock {\em Proc. the First IEEE International Workshop on Higher-Level
  Knowledge in 3D Modeling and Motion Analysis}, pages 58--, 2003.

\bibitem{shapegrammar2006}
Pascal M\"{u}ller, Peter Wonka, Simon Haegler, Andreas Ulmer, and Luc Van~Gool.
\newblock Procedural modeling of buildings.
\newblock {\em ACM Trans. Graph.}, 25:614--623, 2006.

\bibitem{kanji06}
Tanaka K and Kondo E.
\newblock Incremental ransac for online vehicle relocation in large dynamic
  environments.
\newblock {\em Proc. IEEE Int. Conf. Robotics and Automation}, pages
  1025--1030, 2006.

\bibitem{shogo2013partslam}
Hanada Shogo and Tanaka Kanji.
\newblock Partslam: Unsupervised part-based scene modeling for fast succinct
  map matching.
\newblock In {\em IEEE/RSJ Int. Conf. IROS}, pages 1582--1588. IEEE, 2013.

\bibitem{icra15a}
Masatoshi Ando, Yuuto Chokushi, Kanji Tanaka, and Kentaro Yanagihara.
\newblock Leveraging image-based prior in cross-season place recognition.
\newblock In {\em ICRA}, 2015.

\bibitem{ppniv15}
Liu Enfu and Tanaka Kanji.
\newblock Discriminative map matching using view dependent map descriptor.
\newblock In {\em IROS15 WS PPNIV}, 2015.

\bibitem{paz2008large}
Lina~M Paz, Pedro Pini{\'e}s, Juan~D Tard{\'o}s, and Jos{\'e} Neira.
\newblock Large-scale 6-dof slam with stereo-in-hand.
\newblock {\em Robotics, IEEE Transactions on}, 24(5):946--957, 2008.

\bibitem{tanaka2012multi}
Kanji Tanaka and Kensuke Kondo.
\newblock Multi-scale bag-of-features for scalable map retrieval.
\newblock {\em JACIII}, 16(7):793--799, 2012.

\bibitem{MWorg}
James~M Coughlan and Alan~L Yuille.
\newblock Manhattan world: Compass direction from a single image by bayesian
  inference.
\newblock In {\em Computer Vision, 1999. The Proceedings of the Seventh IEEE
  International Conference on}, volume~2, pages 941--947. IEEE, 1999.

\bibitem{thrun2005probabilistic}
Sebastian Thrun, Wolfram Burgard, and Dieter Fox.
\newblock {\em Probabilistic robotics}.
\newblock MIT press, 2005.

\end{thebibliography}

\end{document}